\documentclass[final,3p,times,authoryear]{elsarticle}

\usepackage{amssymb}
\usepackage{amsmath}
\usepackage{graphicx}
\usepackage{subcaption}
\usepackage{booktabs} 

\usepackage{url} 

\journal{arXiv}

\begin{document}

\begin{frontmatter}



\title{Robots and Children that Learn Together : \\ Improving Knowledge Retention by Teaching Peer-Like Interactive Robots} 

\author[label1]{Imene Tarakli \corref{cor1}} 
\ead{i.tarakli@shu.ac.uk}
\author[label1]{Samuele Vinanzi} 
\author[label2]{Richard Moore} 
\author[label1]{Alessandro Di Nuovo} 


\affiliation[label1] {organization={Department of Computing, Sheffield Hallam University},
            city={Sheffield},
            country={United Kingdom}}

\affiliation[label2] {organization={ LOHA Health Ltd},
            city={London},
            country={United Kingdom}}
\begin{abstract}

Despite growing interest in Learning-by-Teaching (LbT), few studies have explored how this paradigm can be implemented with autonomous, peer-like social robots in real classrooms. Most prior work has relied on scripted or Wizard-of-Oz behaviours, limiting our understanding of how real-time, interactive learning can be supported by artificial agents. This study addresses this gap by introducing Interactive Reinforcement Learning (RL) as a cognitive model for teachable social robots. We conducted two between-subject experiments with 58 primary school children, who either taught a robot or practised independently on a tablet while learning French vocabulary (memorisation) and grammatical rules (inference). The robot, powered by Interactive RL, learned from the child’s evaluative feedback. Children in the LbT condition achieved significantly higher retention gains compared to those in the self-practice condition, especially on the grammar task. Learners with lower prior knowledge benefited most from teaching the robot. Behavioural metrics revealed that children adapted their teaching strategies over time and engaged more deeply during inference tasks. This work makes two contributions: (1) it introduces Interactive RL as a pedagogically effective and scalable model for peer-robot learning, and (2) it demonstrates, for the first time, the feasibility of deploying multiple autonomous robots simultaneously in real classrooms. These findings extend theoretical understanding of LbT by showing that social robots can function not only as passive tutees but as adaptive partners that enhance meta-cognitive engagement and long-term learning outcomes.

\end{abstract}





\end{frontmatter}

\section{Introduction}
    \label{sec1}

    Recent advances in Artificial Intelligence (AI) and robotics are reshaping the landscape of education by enabling new forms of interactive and personalised learning. These technologies provide platforms for delivering engaging, student-centred activities tailored to the specific needs and learning paces of children \citep{wang2024development}. Importantly, this technological shift holds the potential to address one of the most pressing challenges in global education: equitable learning opportunities \citep{henkel2025supporting}. By adapting content to individual learners, AI-driven educational tools can support cognitively and academically appropriate curricula, ensuring that children, regardless of their socio-economic background, have access to high-quality education \citep{tu2025empowering,almousa2023conceptualization}.

    Among emerging technologies, social robotics has gained significant attention in educational contexts. Social robots, characterised by their physical presence and capacity for social interaction, have demonstrated promising affective and cognitive benefits for children \citep{belpaeme2018social}. This includes enhanced motivation \citep{donnermann2024integration, song2024impact}, sustained attention \citep{chiang2023improving}, and increased engagement in learning activities \citep{nasir2024social, bruzzo2024charm}. These benefits are particularly salient when considering that learning is inherently a social endeavour, as emphasised by sociocultural theories of education. \cite{vygotsky1978mind} concept of the Zone of Proximal Development underscores the importance of social interaction in cognitive development, suggesting that learners achieve higher levels of understanding through collaborative activities with more capable peers or guides. In this context, social robots can serve as interactive partners that scaffold children's learning experiences, providing immediate feedback and adapting to individual needs, thereby facilitating deeper engagement with the learning materials. 

    One effective way to integrate robots into classrooms is to introduce them as peers, encouraging children to step into the role of a teacher. This method builds on the learning-by-teaching (LbT) paradigm, an educational framework rooted in sociocognitive theories that view learning as a socially mediated process. When children take on the role of a teacher, they engage in deeper cognitive processing by organising, articulating, and evaluating the learning material for a tutee. This promotes meta-cognitive engagement, requiring the tutor to assess the learner’s knowledge and adjust explanations accordingly \citep{duran2017learning}. A central mechanism behind this effect is the Protégé Effect, which suggests that learners put more effort into mastering material when they are responsible for teaching it to someone else \citep{chase2009teachable}. This increased sense of responsibility enhances motivation and fosters deeper learning and reflection \citep{bjork2011making}.
    However, the success of LbT relies not only on the child acting as a tutor, but also on the design of the tutee \citep{serholt2022comparing}. Research shows that for LbT to be effective, the tutee must behave like a novice learner, providing the tutor with opportunities to explain, correct, and reflect \citep{roscoe2008tutor}. The tutee’s behaviour should be carefully calibrated: it must show enough struggle to elicit tutoring behaviours, but also demonstrate gradual improvement, reinforcing the child’s perception of being an effective teacher and sustaining engagement.
    \begin{figure}[]
            \centering
            \includegraphics[width=0.6\textwidth]{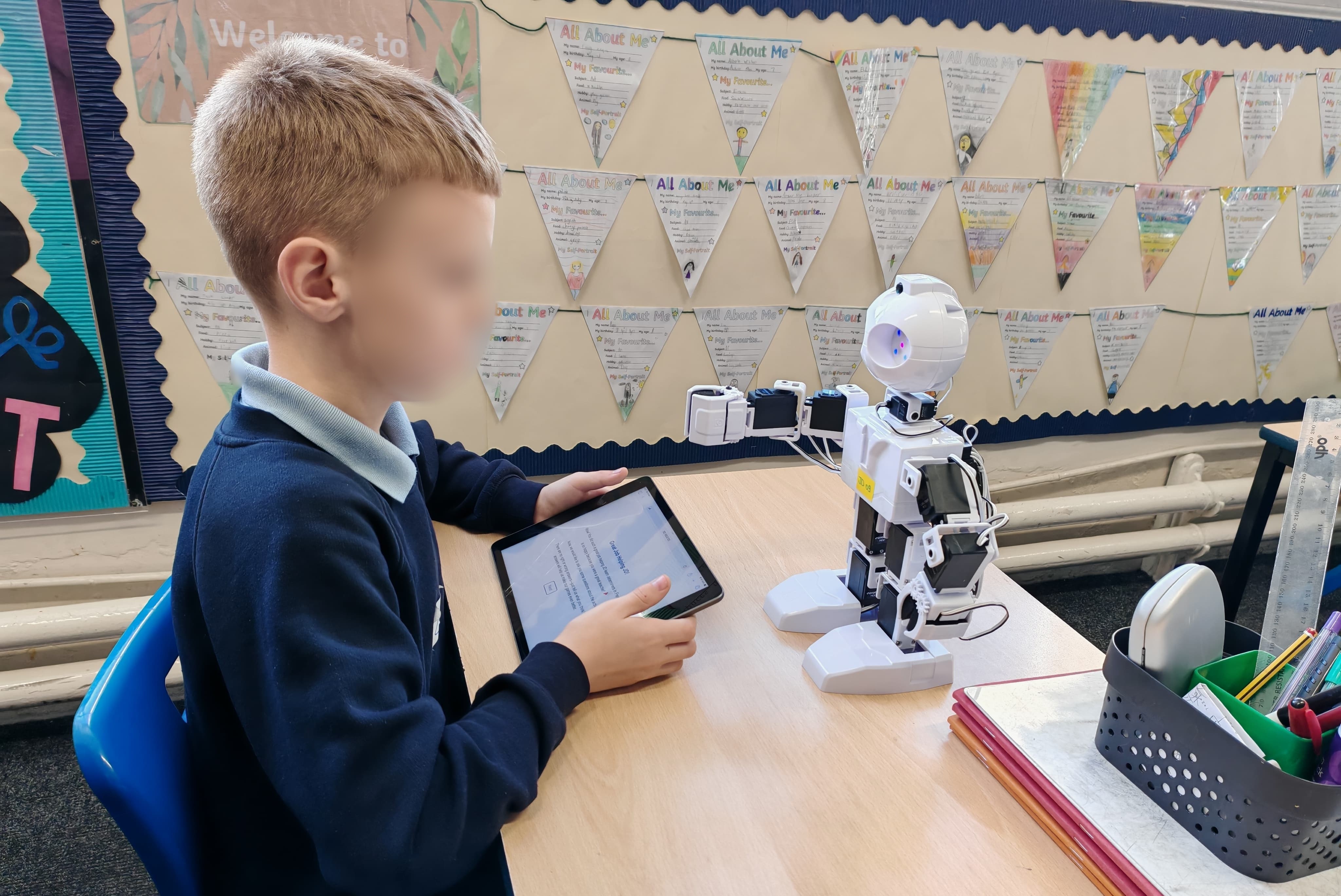}
            \caption{A child interacting with the robot in the Learning-by-Teaching condition. The child provides feedback on the robot’s responses, guiding its learning through evaluative feedback. }
            \label{fig:photo}
        \end{figure}
    Initial implementations of LbT relied on virtual agents and teachable software. These systems allowed children to guide artificial agents by providing examples, demonstrations, or feedback \citep{biswas2005learning, obayashi2000construction}. While effective in controlled settings, these approaches often lacked social presence, limiting their ability to replicate the richness of peer-to-peer learning found in classrooms.
    To address this, researchers introduced social robots as peer-like tutees. Compared to virtual agents, robots provide embodied interaction, enabling richer communication through gaze, gestures, and physical presence; all of which are critical to the social dynamics of teaching \citep{demir2020l2, alimardani2022motivational}. Robots presented as novice peers seeking the child’s help can elicit spontaneous teaching behaviours \citep{tanaka2007socialization}, reduce performance anxiety \citep{lemaignan2016learning}, and foster more natural, bi-directional interaction \citep{pareto2022children, okazaki2015building}. By engaging with a robot peer, children feel less pressure, as they are not being evaluated but instead are offering support; a shift that can be particularly valuable in inclusive or mixed-ability classrooms.
    \\
    Several systems have explored how social robots can embody the tutee role in LbT contexts across a range of domains. Early work by \cite{tanaka2012children} showed that children spontaneously adopted the teacher role when guiding a robot through vocabulary tasks. \cite{lemaignan2016learning} demonstrated how children improved a robot’s handwriting through corrective feedback, triggering meta-cognitive engagement. \cite{yadollahi2018deictic} applied the LbT paradigm to reading, where children helped the robot overcome errors in word pronunciation. Expanding to group settings, \cite{el2019learning} investigated collaborative teaching of a robot in a handwriting task, showing that group-level reflection supported meta-cognitive engagement, regardless of the engagement prompt used. Similarly,  \cite{verhoeven2018designing} developed a story-driven language learning activity in which children taught a robot new vocabulary through playful storytelling interactions. \cite{pareto2022children} explored how children teach a robot or a younger peer using a structured tutoring system (SPARk), showing that robot tutees can elicit similar teaching behaviours to those seen in peer-to-peer interactions. More recently, \cite{chen2024impact} implemented an LbT approach using an AI-powered robot for image recognition in biology lessons. Students who taught the robot to recognise cell division stages showed significantly higher conceptual understanding and motivation compared to textbook learners.

    To support such interactions, various models have been developed to simulate novice-like behaviour in social robots. These models aim to present the robot as a peer who is learning from the child, thereby encouraging tutoring behaviours and cognitive engagement. Across the literature, three primary approaches can be identified. Wizard-of-Oz (WoZ) models rely on a hidden human operator to control the robot’s responses, allowing for smooth interaction but offering no real autonomy or scalability \citep{das-pon-barry-2018-turn, 10.1007/978-3-030-06134-0_42}. Rule-based models script the robot’s behaviour in advance, enabling it to simulate learning progression, but these are rigid and cannot adapt to individual learners or dynamic learning conditions \citep{serholt2022comparing, lee2021curiosity, el2019learning}. Learning-based models offer a more promising alternative by allowing the robot to update its behaviour over time, better emulating a genuine learning process \citep{chen2024impact, pareto2022children,chandra2020children}. However, most of these systems are highly task-specific or depend on pretraining rather than real-time adaptation. For a robot to truly act as a peer learner, it must not only behave like a novice but also improve in ways that reflect the child’s input, reinforcing the child’s sense of responsibility and sustaining engagement over time. This requires the robot to follow an emergent learning trajectory that evolves throughout the interaction.

    To meet this need, we propose the use of teachable robots; robots that can learn directly from the child’s guidance. In particular, we focus on Interactive Reinforcement Learning, a paradigm in which the robot receives evaluative feedback from the user (e.g., whether an answer was correct or wrong) and updates its policy accordingly. Unlike traditional RL, Interactive RL is designed to work with non-expert users and is thus well suited to classroom environments, enabling the robot to learn in real time while also encouraging the child to reflect on and monitor the robot’s learning process. Pedagogically, this approach offers several advantages aligned with current trends in computer-supported education: it provides a form of adaptive scaffolding, where the robot’s learning path is shaped by the student’s input; it reinforces formative assessment, as the child continually evaluates and corrects the robot’s performance; and it promotes metacognitive development by prompting learners to reflect not only on what they know, but on how to teach and explain it. By integrating Interactive RL into the robot’s cognitive model, we aim to create a socially engaging and pedagogically meaningful tutee; one that supports personalisation, deepens learning, and encourages active knowledge construction in line with contemporary educational goals.

    Building on these foundations, our work investigates how children can teach a social robot in real time using evaluative feedback, and how this interaction affects their learning and engagement. By integrating Interactive RL into a teachable peer robot, we aim to examine not only the technical feasibility of such systems, but also their pedagogical potential in classroom settings. Specifically, we address the following research questions:
    \begin{itemize}
        
        \item \textbf{RQ1: Can primary school children effectively guide a social robot’s learning using Interactive Reinforcement Learning in a real classroom environment?} \\
        This question explores the feasibility of integrating Interactive RL in naturalistic educational settings, assessing whether children are capable of providing accurate, meaningful feedback that helps the robot learn.
        
        \item \textbf{RQ2: Does teaching a robot improve children’s long-term knowledge retention compared to independent practice?} \\
        Here, we examine whether learning-by-teaching a robot leads to better consolidation of knowledge than more traditional self-practice methods, reflecting pedagogical goals related to memory, transfer, and depth of processing.

        \item \textbf{RQ3: Do children with lower prior knowledge benefit more from the learning-by-teaching interaction than their higher-achieving peers?} \\
        This question addresses issues of educational equity and personalization by investigating whether teachable robots can provide greater scaffolding for learners who might otherwise struggle in traditional environments.

        \item \textbf{RQ4: How does the nature of the learning task influence the effectiveness of teaching a robot?} \\
        This question explores how different types of cognitive processing—recall versus conceptual reasoning—shape the outcomes of robot-based LbT interactions, with implications for task design and adaptive learning systems.
    \end{itemize}

    The remainder of the paper is structured as follows. We first describe the methodology, detailing the design of the cognitive model, the learning tasks, and the experimental setup implemented in a real classroom setting. We then present the results, examining the effects of the LbT interaction on children's performance, engagement, and retention across different tasks and learner profiles. This is followed by a discussion of the findings in relation to existing literature, highlighting both pedagogical implications and design considerations for teachable robots. Finally, we conclude by summarising our contributions, discussing limitations, and outlining directions for future research.

\section{Method}  

To investigate the research questions, we conducted a classroom-based study in which primary school children taught a social robot using evaluative feedback. This section describes the cognitive model underlying the robot’s behaviour, the learning tasks implemented, the experimental design, and the evaluation measures used to assess learning outcomes and interaction quality. 

    \subsection{Research Model and Procedure}
    
\begin{figure}[h]
        \centering
        \includegraphics[width=0.9\textwidth]{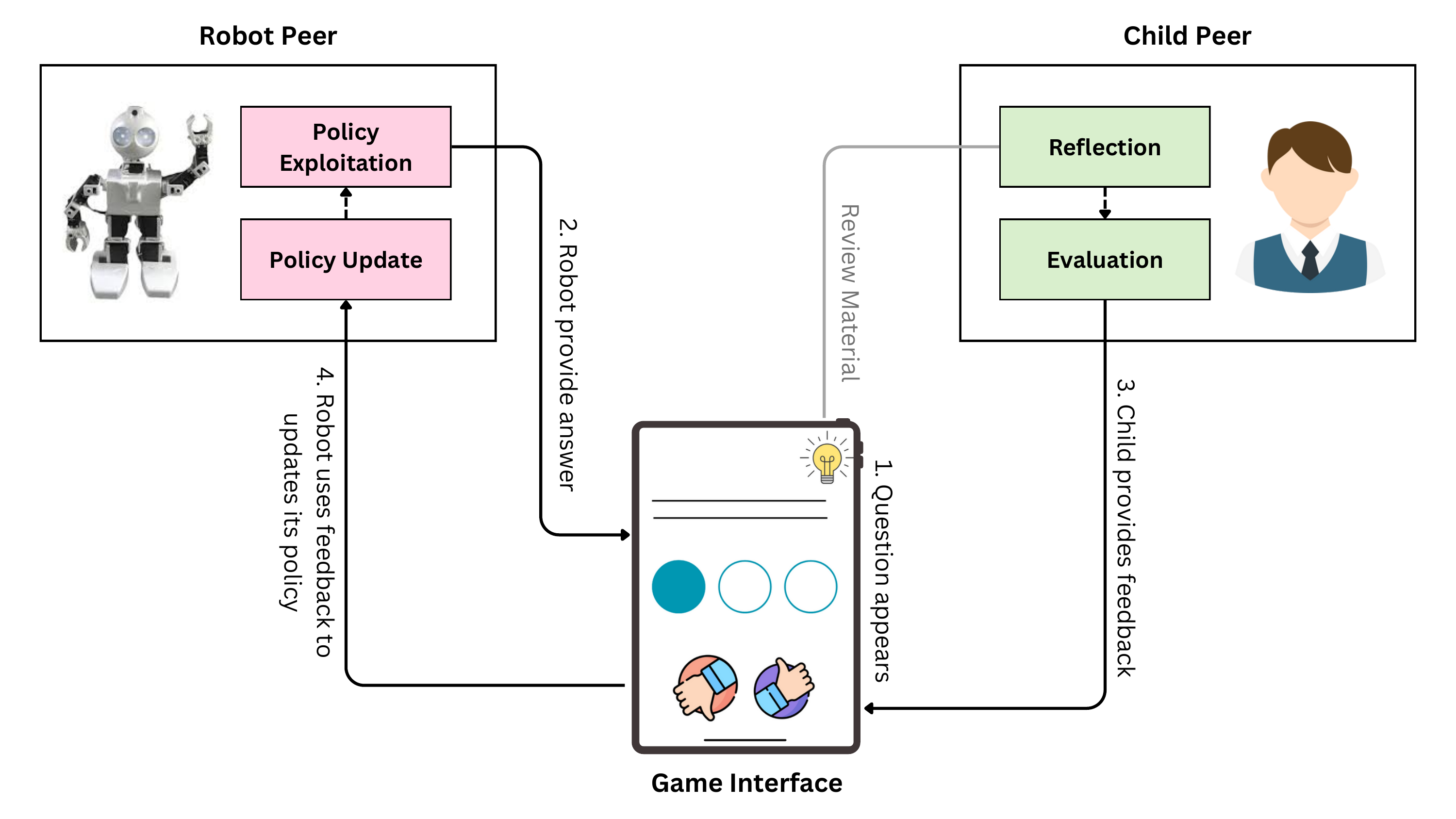}
        \caption{ Overview of the Learning-by-Teaching framework with Interactive Reinforcement Learning. (1) A question is displayed with multiple choice options.(2) Policy Exploitation: The robot peer uses its current policy  to select an answer. (3) The child peer reflects on the robot’s answer (Reflection), optionally reviews the material, and decides whether the robot’s response is correct or incorrect (Evaluation).(4) The robot updates its policy (Policy Update) based on the child’s feedback. This loop continues as the robot progressively refines its knowledge, while the child simultaneously reinforces their own understanding through active teaching. }
        \label{fig:framework}
    \end{figure}

    We designed a cognitive framework that allows the peer robot to emulate a child’s learning trajectory using Interactive RL. This approach was implemented within a game-based educational activity, where the robot begins as a novice learner, selecting answers at random. As the child tutor provides evaluative feedback, the robot incrementally refines its policy, progressively improving its performance across the learning session. Figure~\ref{fig:framework} illustrates the overall structure of the learning-by-teaching framework.

        \subsubsection{Task game and policy formulation}
        Game-based learning has been widely recognised for its positive impact on children’s learning outcomes \citep{magpusao2024gamification}. Building on this, we designed an educational quiz game in which the robot, acting as a learner, selects answers from multiple options and receives evaluative feedback from the child. Unlike traditional quiz formats, the child does not answer directly but instead assumes the role of a tutor, guiding the robot’s learning by indicating whether its responses are correct or incorrect.
        
        We formalise this interaction using a Markov Decision Process (MDP), where each question represents a state \( S = \{ \text{question}_1, \dots, \text{question}_n \} \), and the available answer options define the action space \( A = \{ \text{option}_1, \dots, \text{option}_m \} \). The robot's policy reflects its evolving knowledge representation; an optimal policy maps each question to the correct answer, representing the desired learning outcome.
        
        To ensure the pedagogical relevance of the game, the learning tasks were selected in collaboration with primary school teachers. The chosen activities had to involve content not previously covered in class, minimise ceiling effects, and present an appropriate cognitive challenge. Tasks that are too simple may result in disengagement \citep{tanaka2012children}, while more conceptually demanding activities are known to stimulate learning and reflection \citep{Authors2023, chen2020teaching}. Based on these criteria, we implemented two tasks: a vocabulary memorisation task on French body parts and a grammar rule inference task on French determiners.
        
        \paragraph{Body Parts Game} This task focuses on teaching French vocabulary by matching English body part names (\textit{hand, head, foot, belly,} and \textit{eye}) to their French equivalents. Each round displays one English word and three possible translations, from which the robot selects an answer. This task leverages prior findings on the effectiveness of social robots in language learning, particularly when iconic gestures are used to ground word meanings \citep{de2018effect}. During the interaction, the robot reinforces feedback by pointing to the relevant body part. This task is modelled as an MDP with a state space of five (targeted words) and an action space of three (translation options).
        
        \paragraph{Grammar Game} This task targets rule inference by asking the robot to classify a French word into one of three grammatical categories (feminine, masculine, or plural) based on its determiner (\textit{La, Le, Un, Une, Les,} or \textit{Des}). The child assesses the robot’s choice, reinforcing or correcting it. Prior work suggests that learning-by-teaching is especially effective in tasks that involve conceptual reasoning and meta-cognitive engagement \citep{biswas2005learning}. This task is modelled as an MDP with a state space of six (determiners) and an action space of three (grammatical categories). \\

        The quiz game was developed as a web-based interface using Flask and deployed on touchscreen tablets. Each game question corresponded to a state in the MDP, and each answer choice mapped to a possible action. This interface served as the medium for both the robot's decision-making and the child’s feedback, enabling real-time interaction and policy updates.

            \subsubsection{Interactive Reinforcement Learning Peer Model}
    
            RL is a computational approach used to solve MDPs by enabling an agent to learn optimal decision-making strategies through trial and error. A reward function 
            \(R: S \times A \to \mathbb{R}\) assigns a numerical value to each state-action pair, reflecting the immediate benefit of taking a specific action in a given state. The agent aims to discover an optimal policy \(\pi^*\) that maximises the expected cumulative reward over time by selecting actions that maximise the Q-value, an estimate of the long-term return:
            \begin{equation}
                Q(s,a) = \mathbb{E}_{\pi} \left[\sum_{t=0}^{\infty} \gamma^t R(s_t, a_t)\right].
            \end{equation}
            
            Interactive RL extends this framework by allowing non-expert users to guide the agent through evaluative feedback rather than relying on a predefined reward function \citep{najar2021reinforcement}. This human-in-the-loop approach is particularly well suited to LbT scenarios, where children serve as the source of feedback, shaping the robot’s learning trajectory in real time.
            
            In our setup, the peer robot uses the TAMER framework \citep{knox2009interactively}, which interprets binary feedback as immediate evaluative input. After each action, the robot receives a signal from the child indicating whether the response was correct (+1) or incorrect (-1). The Q-value is updated accordingly using the following rule:
            \begin{equation}
                Q'(s, a) = Q(s, a) + \alpha \cdot (h - Q(s, a))
            \end{equation}
            where \(Q'(s, a)\) is the updated value of the state-action pair, \(Q(s, a)\) is the previous estimate, \(h\) is the feedback signal, and \(\alpha\) is the learning rate.
            
            This interactive feedback loop enables the robot to progressively refine its policy based on the child’s input, allowing it to emulate a meaningful and adaptive learning trajectory. In doing so, the robot not only improves its own performance but also fosters reflective thinking and engagement in the child acting as the tutor.

            \subsubsection{Robot-Child Interaction and Experimental Setup}
    
            The learning scenario was deployed using a hybrid system comprising three components: (1) the JD social robot, (2) a touchscreen tablet running the learning game, and (3) a central computer hub that controlled the robot’s cognitive model and managed synchronisation via MQTT communication. This setup enabled real-time interaction between the child and the robot, allowing the robot to receive feedback and update its behaviour accordingly.

            The robot used in this study, JD\footnote{\url{https://www.ez-robot.com}}, is a compact humanoid platform with articulated limbs and LED-animated eyes, designed to be visually engaging and accessible for young learners. To make the robot more appealing to children, its default voice was replaced with a natural, child-like voice generated using ElevenLabs\footnote{\url{https://elevenlabs.io/}}. Although primarily intended for programming education, JD was repurposed as a peer-like tutee for this study, with custom behaviours developed to support naturalistic interaction, including speech-aligned head movements and gestures.
    
             Interaction took place via a touchscreen tablet, which served as the primary medium for presenting questions and recording child feedback. Each screen displayed a question and answer choices, along with a progress tracker and hint button. After the robot selected an answer, feedback buttons appeared, enabling the child to indicate whether the response was correct or incorrect.
            
            Each game round followed a structured interaction sequence. The robot introduced itself and the learning task, then attempted to answer each question. It waited for the child’s feedback, prompting if no response was given within ten seconds. If no feedback was received within the next fifteen seconds, the robot invited the child to use the hint button.  Feedback was acknowledged both verbally and visually, with eye colour changes (green for correct, red for incorrect). When incorrect, the robot and child reviewed the correct answer together. In the vocabulary task, the robot also used iconic gestures, such as pointing to the relevant body part, to reinforce the association. At the end of the session, the robot thanked the child and simulated “going to sleep” to signal the conclusion of the activity.
            
            To preserve ecological validity, the experiment was conducted in a real classroom setting, with children remaining in their regular groups. Multiple one-on-one child-robot interactions were carried out simultaneously within the same classroom to reflect a natural learning environment. This setup also minimised direct experimenter interaction, reducing the risk of social desirability bias and suggestibility effect often observed in child studies \citep{zaga2015effect}. Maintaining the peer group context was especially important given the study’s focus on peer-like robot interaction and collaborative learning dynamics.
            
    \subsection{Research Context and Sample}
    The study was conducted in collaboration with a UK primary school to evaluate the effectiveness of LbT interaction in a real-world classroom environment. In what follows, we describe the participant group and outline the study protocol followed. 

        \subsubsection{Participants}
        A total of 58 children aged 8–9 years (Year 4, UK) participated in the study. All children attended the same primary school in the UK, were distributed across two classes, and studied French as an additional language. Parental consent was obtained prior to the study, along with verbal assent from the children on the day of participation. To ensure data privacy, we pseudonymised the data using a system that combined each child’s classroom number with the first letter of their teacher’s name. This approach allowed us to match retention scores with data collected during the study while maintaining confidentiality.
        Since LbT is intended for knowledge consolidation rather than acquiring new knowledge, we excluded one participant who scored 0 on the pre-test. Additionally, due to data logging errors, technical issues, and instances where children entered incorrect pseudonymisation IDs, the final dataset included 53 participants (33 boys, 16 girls, and 4 who preferred not to specify their gender).

        \subsubsection{Study Protocol}

\begin{figure}[]
            \centering
            \includegraphics[width=0.6\textwidth]{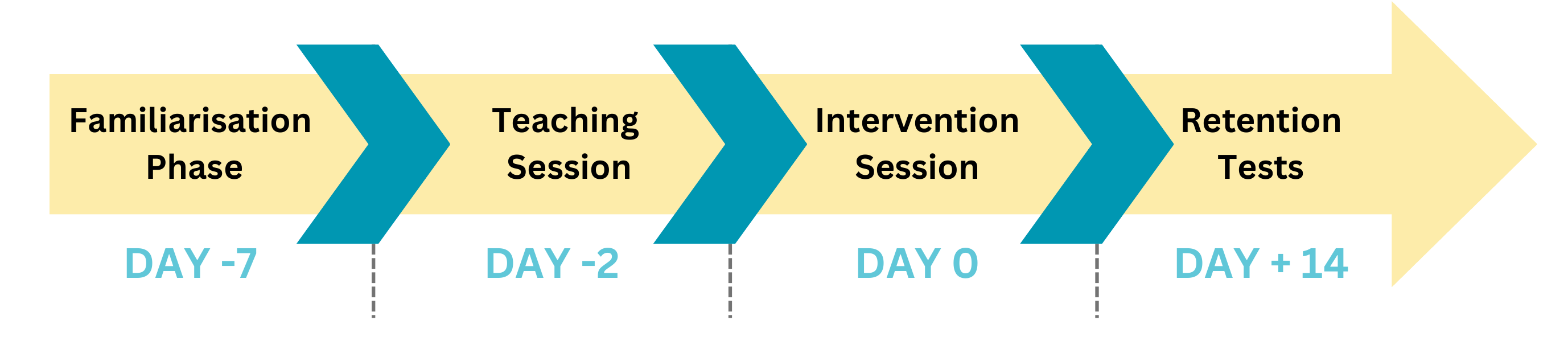}
            \caption{Study timeline. Familiarisation Phase (Day -7) to introduce children to the social robot, the Teaching Session (Day -2) where students learn the material, the Intervention Session (Day 0) where they engage in either Learning-by-Teaching or Self-Practice, and the Retention Test (Day +14) to assess long-term knowledge retention. }
            \label{fig:proto}
        \end{figure}

        The study was conducted over a three-week period and consisted of four phases: (1) a familiarisation session, (2) a classroom teaching session led by the teacher, (3) the main intervention session involving the robot or tablet activity, and (4) a delayed retention test conducted two weeks later, as illustrated in Figure~\ref{fig:proto}.

        \paragraph{Familiarisation Session}  
        To reduce novelty effects and ensure children were comfortable interacting with the robot, a familiarisation session was conducted one week prior to the intervention. The JD robot was introduced in the classroom, where it greeted the children, demonstrated basic movements, and performed a short dance. To help children practice providing feedback, a short group activity was carried out on the classroom touchboard, where children took turns teaching the robot the names of fruits in French using the same feedback buttons they would later use during the study. The session concluded with a Q\&A where children could ask questions about the robot’s capabilities.
        
        \paragraph{Teaching Session}  
        Two days before the intervention, teachers conducted standardised lessons on French body parts and determiners using materials provided by the researchers. This session ensured that all children had a basic level of prior knowledge and positioned the robot as a learning companion rather than a replacement for instruction. The timing of this session was chosen to balance memory retention and prevent ceiling effects during pre-testing.
        
\begin{figure}[]
                \centering
                \includegraphics[width=0.85\textwidth]{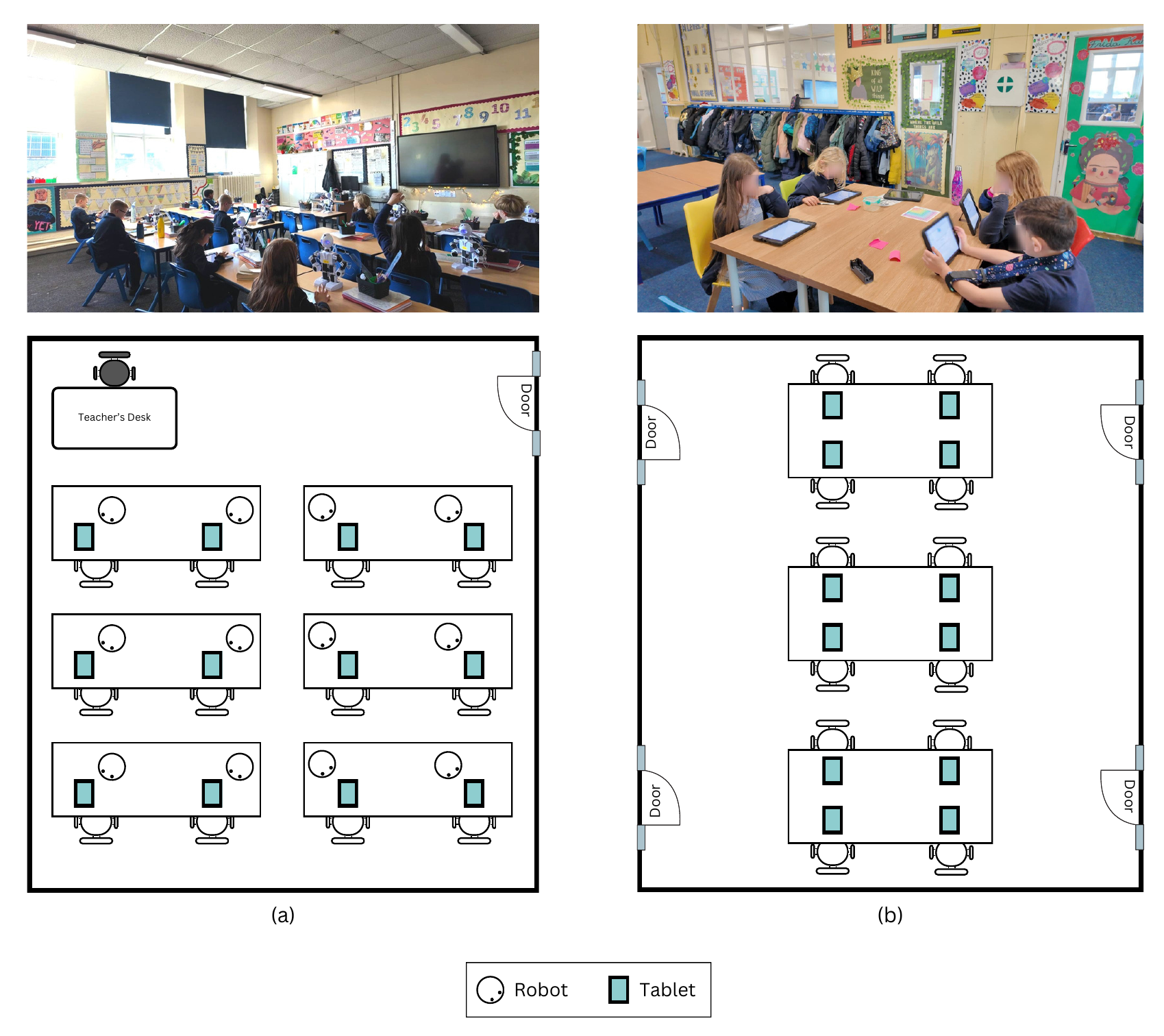}
                \caption{Floor plan of the study setup for both conditions. (a) Learning by Teaching condition: Children interacted with the robot in a classroom setting, with robots and tablets distributed around the room. (b) Self-Practice condition: Children worked independently on tablets in a communal area. }
                \label{fig:floor_plan}
            \end{figure}

        \paragraph{Intervention Session}  
        The study was structured as two separate between-subject experiments; one for each learning task (Body Parts and Grammar). This decision was made to avoid potential carryover effects and ensure that learning gains could be attributed to the specific learning condition rather than task repetition or familiarity. A within-subject design was not feasible, as it would have required children to complete both conditions on the same task, which could confound learning effects and increase the risk of ceiling effects in post-tests. Moreover, to prevent social exclusion and maintain classroom cohesion, we ensured that all children experienced both robot and tablet conditions, but on different tasks.
        Each class was divided into three groups: one group interacted with the robot, another completed the same task independently on a tablet (control condition), and the third group participated in an unrelated computer science activity. Groups rotated through the activities to ensure all children experienced both learning conditions, but on different tasks. Specifically, children who taught the robot in the Body Parts game used the tablet for the Grammar game, and vice versa.
        Each session lasted approximately 20 minutes and began with pre-tests to assess prior knowledge. Children then engaged in the learning game for 15 iterations, followed by immediate post-tests and a brief questionnaire. The robot sessions took place in the classroom, while tablet sessions were held in an adjacent communal space. This structure preserved a naturalistic learning environment while enabling controlled comparisons. Figure~\ref{fig:floor_plan} illustrates the floor setting of the study. 
        
        \paragraph{Retention Test}  
        Two weeks after the intervention, children completed the same learning tests again to assess long-term retention. These tests were conducted on tablets during regular class hours. Children who had participated in both the robot and tablet conditions were tested on both tasks. As a gesture of appreciation, children received certificates of participation, and the JD robot performed a dance at the end of the session.

    \subsection{Instrument Used and Their Validation}
   
    To assess the effects of the learning-by-teaching interaction, we employed three types of instruments: (1) knowledge tests, (2) in-game performance metrics, and (3) perception questionnaires. These instruments were designed to evaluate learning outcomes, engagement, and interaction quality across conditions.
    
    \paragraph{Knowledge Tests}
    Children completed three touchscreen-based tests for each task: a pre-test (to assess prior knowledge), a post-test (immediately after the intervention), and a delayed retention test (two weeks later). All tests were implemented on tablets using the same interface as the learning game to ensure consistency. Each test included three rounds of activities to reduce random guessing.
    
    For the \textit{Body Parts} task, tests included: (1) matching English body part names to their French equivalents, (2) matching French words to English equivalents, and (3) matching French words to images. These tasks assessed memorisation.
    
    For the \textit{Grammar} task, the test required children to classify nouns as masculine, feminine, or plural based on their determiners. To ensure the test measured rule generalisation rather than memorisation, the test items differed from those used during the learning session.
    
    We calculated knowledge gain and retention gain by subtracting the pre-test score from the post-test and retention test scores, respectively. 
    
    \paragraph{In-Game Performance Metrics}
    During the learning activity, we recorded app-based behavioural data to assess real-time engagement and learning strategy use. Logged metrics included:
    
    \begin{itemize}
        \item \textbf{Feedback Accuracy:} Whether the child correctly identified the robot's answers as correct or incorrect.
        \item \textbf{Time per Iteration:} The time spent on each question, from the robot's answer to the child's feedback.
        \item \textbf{Hint Usage:} Whether and how often the child used the hint button to review content before giving feedback.
    \end{itemize}
    
    These measures were used to infer help-seeking behaviour, reflection, and temporal engagement across conditions.
    
    \paragraph{Perception Questionnaire}
    Children completed a brief post-intervention questionnaire to report their perceptions of the learning experience. Items measured task enjoyment, perceived competence, and engagement, and were adapted from the Intrinsic Motivation Inventory (IMI) \citep{ryan1983relation} to suit the age group and study context. Additionally, a single-item question was included to assess perceived task difficulty. 

    To ensure accessibility and reduce cognitive load, all items used a 5-point Likert scale presented as a visual star-rating system. Items were displayed one at a time in random order, and children selected their response by tapping the appropriate number of stars. The adaptation was informed by prior work on child-friendly self-report instruments and was previously tested in a classroom setting to ensure clarity and usability \citep{chandra2020children}. Internal consistency coefficients (Cronbach’s alpha) for each subscale are reported in the supplementary material.
    
    \subsection{Data Analysis}
    Quantitative data from knowledge tests, in-game interaction logs, and questionnaires were analysed using a combination of parametric and non-parametric tests, depending on data distribution. Normality was assessed using Shapiro–Wilk tests, and appropriate comparisons were made using independent-samples \textit{t}-tests or Mann–Whitney \textit{U} tests, with effect sizes reported as Cohen’s \textit{d} or rank-based \textit{r}.
    
    \paragraph{Learning Outcomes}
    Knowledge gain and retention gain were calculated as the difference between post-test or retention test scores and pre-test scores. Between-condition comparisons were conducted for each task, and median-split analyses were performed to explore the impact of prior knowledge (low- vs. high-baseline groups).
    
    \paragraph{In-Game Performance Metrics}
    Metrics including iteration time, hint usage, and feedback accuracy were analysed to assess engagement and reflection. Within-condition progress over time (e.g., first vs. last five iterations) was evaluated using Wilcoxon signed-rank tests, and between-condition differences were assessed using Mann–Whitney \textit{U} tests. Spearman’s rank correlations were computed to explore relationships between behavioural metrics, learning outcomes, and perceived engagement.
    
    \paragraph{Questionnaire Responses}
    Post-intervention questionnaire scores were compared across conditions using Mann–Whitney \textit{U} tests. Exploratory correlations were run with behavioural and performance data. Cronbach’s alpha coefficients for each scale are reported in the supplementary material to confirm reliability of the adapted measures.

\section{Results}

    \subsection{Effectiveness of Teaching Through Feedback}

    We evaluated the feasibility of using Interactive RL as a cognitive model for LbT a social robot, examining whether children could effectively use it to improve their learning and engage in meta-cognitive processes that support knowledge retention.

    Children successfully assumed the role of a teacher, providing feedback throughout the game. The accuracy of the feedback was $0.89 \pm 0.13$ for the Body Parts Game and $0.74 \pm 0.29$ for the Grammar Game, indicating that children were generally able to guide the robot’s learning process effectively.

\begin{figure}[h]
        \centering
        \begin{subfigure}{0.48\textwidth}
            \centering
            \includegraphics[width=1\textwidth]{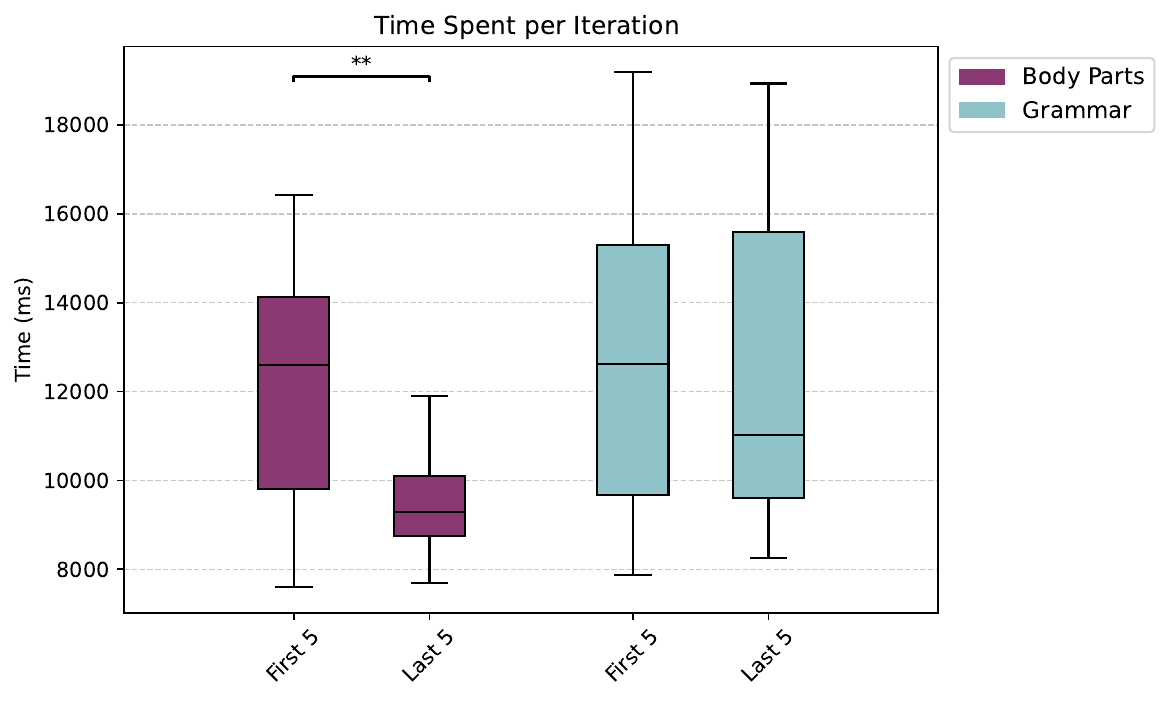}
            \caption{}
            \label{fig:H1_it}
        \end{subfigure}
        \begin{subfigure}{0.48\textwidth}
            \centering
            \includegraphics[width=1\textwidth]{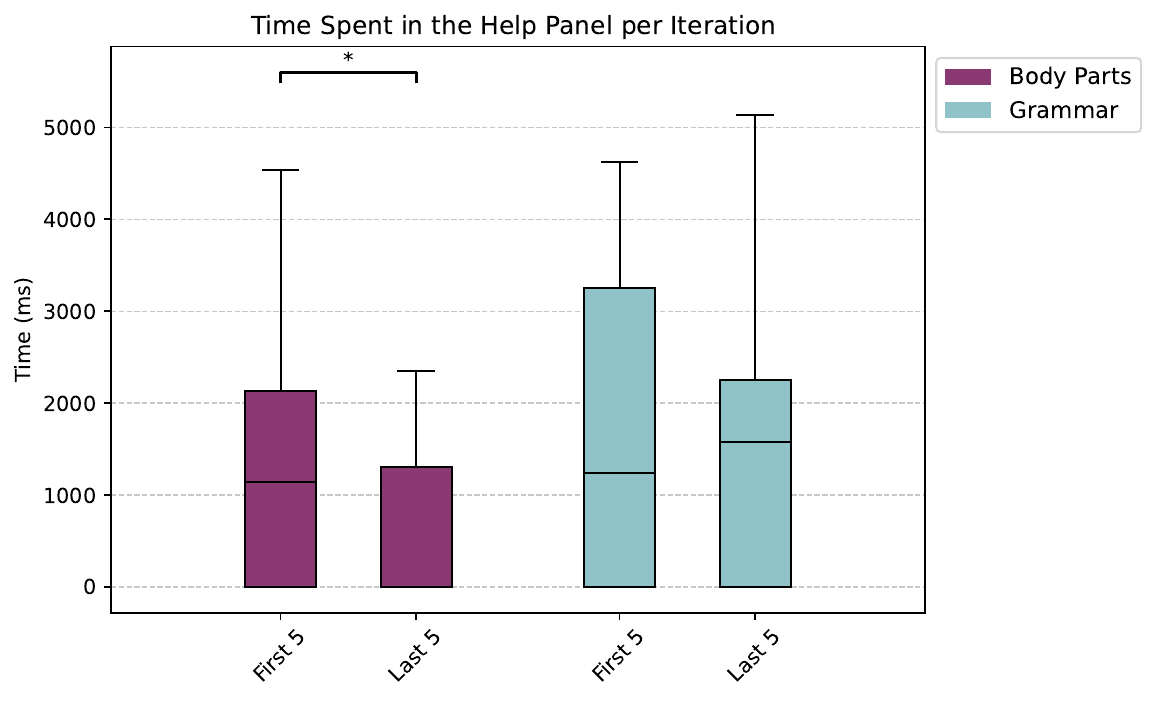}
            \caption{}
            \label{fig:H1_help}
        \end{subfigure}
    
        \caption{Comparison of time spent during the game between the first and last five iterations. (a) Boxplots showing the time spent per iteration. (b) Boxplots illustrating the time spent on the help panel per iteration.}
        \label{fig:H1_time}
    \end{figure}

    Additionally, Figure~\ref{fig:H1_it} shows that children demonstrated adaptation to their role as a teacher by progressively reducing their response time per iteration. This decrease was significant in the Body Parts Game, where children spent less time per iteration in the final five rounds ($Mdn = 9289.20ms$) compared to the first five ($Mdn = 12596.52ms$), $U = 607.00, z = 2.90, p < 0.01, r = 0.38.$
    
    Furthermore, as seen in Figure~\ref{fig:H1_help}, children in the Body Parts Game also spent less time on the help panel towards the end of the game ($Mdn = 0.00ms$) compared to the beginning ($Mdn = 1142.40ms$), $U = 563.50, z = -2.61, p < 0.05, r = -0.34$. However, this change was not significant for the Grammar Game.

    These findings suggest that children were not only able to provide feedback but also demonstrated learning by adjusting their interaction patterns, particularly in the Body Parts Game.
    
    \subsection{Learning by Teaching Vs Self-practice}

    We compared the learning outcomes of learning-by-teaching a social robot with the self-practice condition, while also examining whether the perception of the task differed between the two approaches.

\begin{figure}[h]
    \centering
    \begin{subfigure}{0.48\textwidth}
        \centering
        \includegraphics[width=1\textwidth]{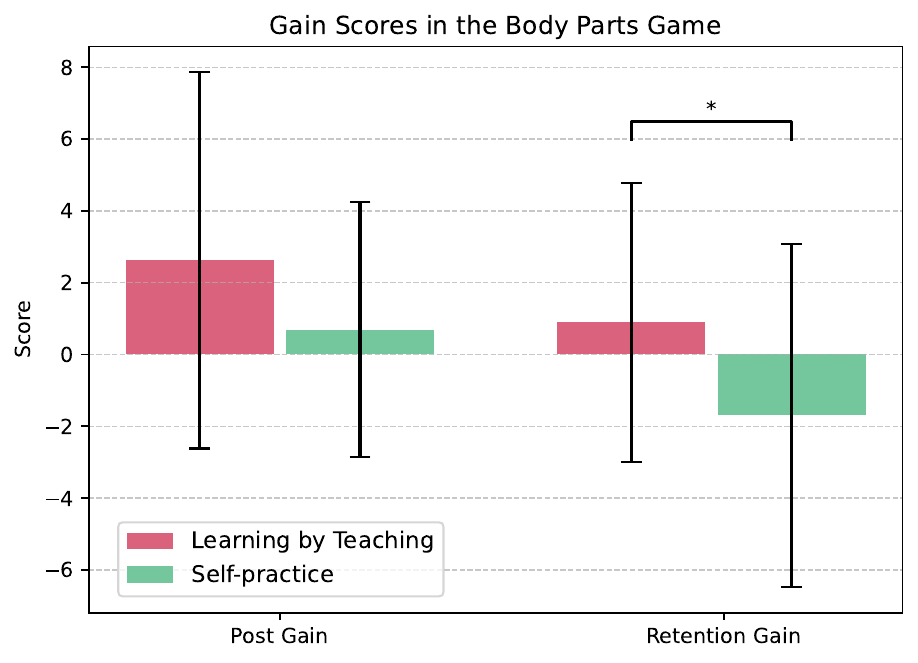}
        \caption{}
        \label{fig:H2_gain_body}
    \end{subfigure}
    \begin{subfigure}{0.48\textwidth}
        \centering
        \includegraphics[width=1\textwidth]{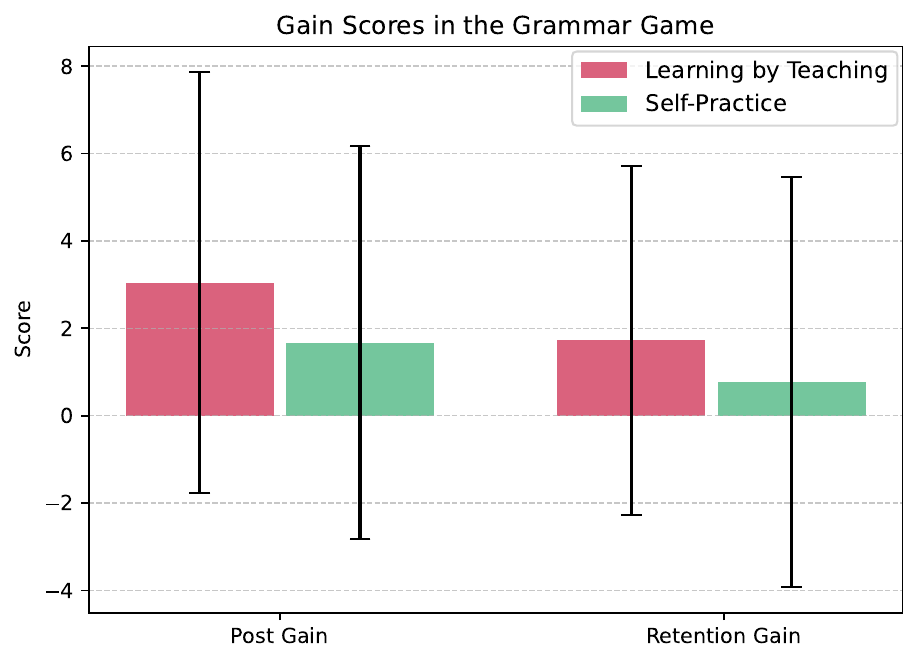}
        \caption{}
        \label{fig:H2_gain_gender}
    \end{subfigure}
    
    \caption{Comparison of learning gains between the Learning-by-Teaching and Self-Practice conditions for the Body Parts Game and Grammar Game. (a) Body Parts Game: Participants in the LbT condition showed significantly higher retention gains compared to the Self-Practice condition ($p < 0.05$). (b) Grammar Game: No significant difference was observed between the two conditions. }
    \label{fig:H2_gain}
    \end{figure}

    Figure~\ref{fig:H2_gain} shows that, on average, participants in the LbT condition achieved higher post- and retention gains compared to those in the Self-Practice condition. This difference was significant in the Body Parts Game, where participants in the LbT condition ($M=0.90, SE =3.89$) demonstrated a higher retention gain than those in the Self-Practice condition ($M=-1.70, SE = 4.78$), $t(50) = 2.16, p<0.05, r=0.29$. 
    
\begin{figure}[h]
    \centering
    \begin{subfigure}{0.48\textwidth}
        \centering
        \includegraphics[width=1\textwidth]{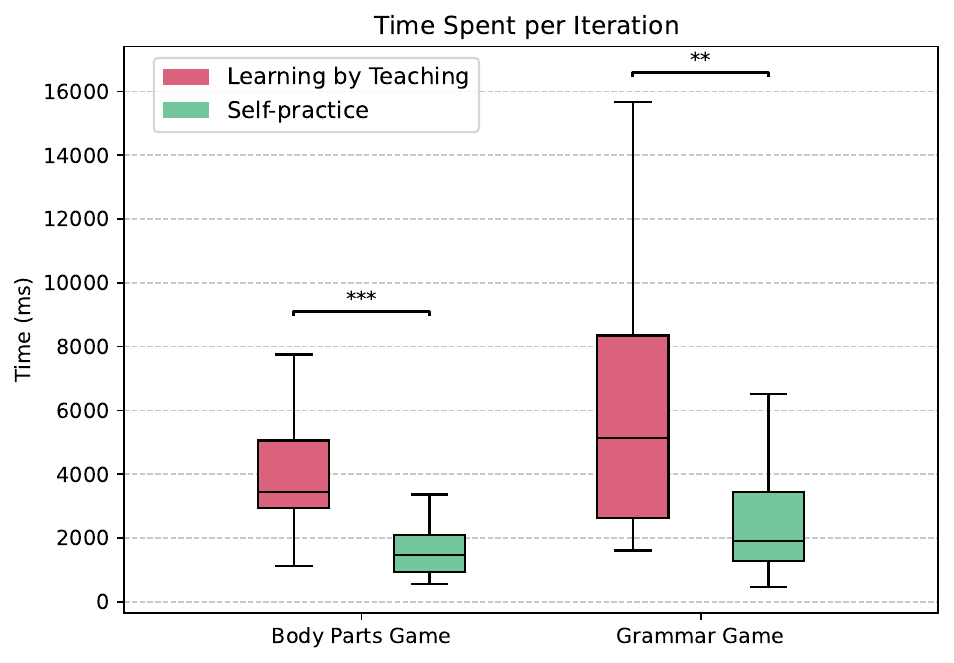}
        \caption{}
        \label{fig:H2_time_it}
    \end{subfigure}
    \begin{subfigure}{0.48\textwidth}
        \centering
        \includegraphics[width=1\textwidth]{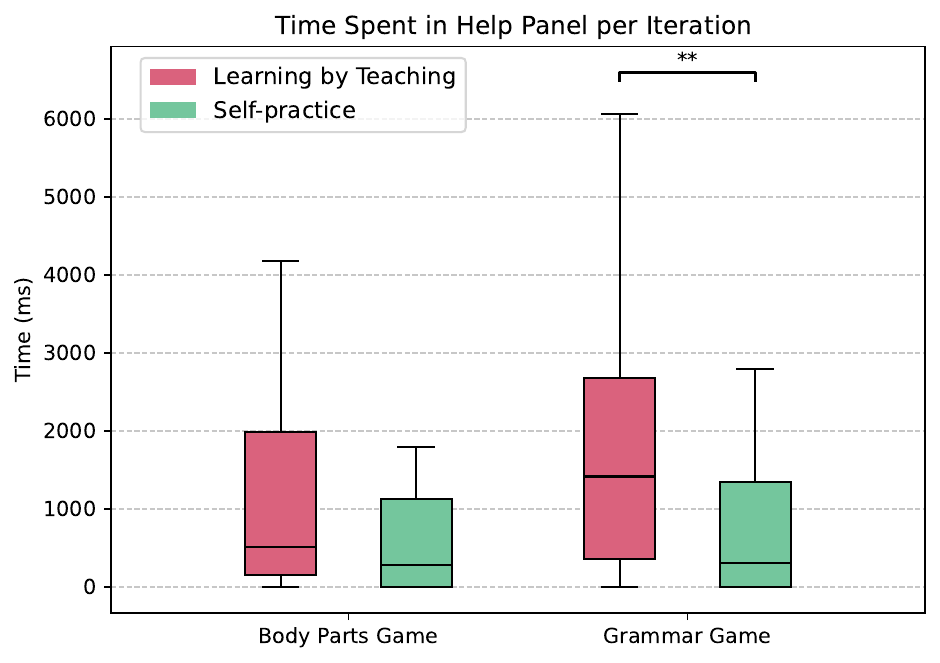}
        \caption{}
        \label{fig:H2_time_help}
    \end{subfigure}
    
    \caption{Comparison of time spent during the game between the Learning by Teaching and Self-Practice conditions for the Body Parts Game and Grammar Game. (a) Time spent per iteration: Children in the LbT condition spent significantly more time per iteration across both tasks ($p < 0.01$). (b) Time spent in the help panel: A significant difference was observed in the Body Parts Game ($p < 0.01$), while no significant difference was found for the Grammar Game.}
    \label{fig:H2_time}
    \end{figure}

    Additionally, participants in the Learning-by-Teaching condition spent significantly more time per iteration compared to those in the Self-Practice condition, as depicted in Figure~\ref{fig:H2_time_it}. Specifically, in the Body Parts Game, children in the LbT condition ($Mdn = 3431.66 ms$) spent more than twice the time per iteration compared to those in the Self-Practice condition ($Mdn = 1476.40ms$), $U=571.0, z=0.88, p<0.0001, r=0.12$.  Similarly, in the Grammar Game, children in the LbT condition ($Mdn = 5127.50ms$) spent significantly more time per iteration than those in the Self-Practice condition ($Mdn= 1916.03ms$),  $U=535.0, z=0.76, p<0.001, r=0.11$. These results indicate that children were more engaged with the learning material in the LbT condition.

    Moreover, Figure~\ref{fig:H2_time_help} shows that children in the LbT condition also spent more time using the help panel than those in the Self-Practice condition. This effect was significant in the Grammar Game, where children in the Learning-by-teaching condition ($Mdn=1417.13ms$) spent four times more time reviewing the material compared to those in the Self-Practice condition ($Mdn=312.80ms$), $U=468.0, z=2.56, p<0.01, r=0.35$. 
    
    Time spent on the help panel in the LbT condition was significantly correlated with retention gain ($r=0.30, p < 0.05$), whereas no such relationship was observed in the Self-Practice condition ($r = -0.01, p > 0.05$). This suggests that children in the LbT condition engaged more actively in self-reflection and material review, which may have contributed to their higher retention gain.

    Table~\ref{table:game_comparison} presents the comparison of perceived engagement, competence, enjoyment, and task difficulty between the LbT and Self-Practice conditions for both learning tasks. Across all measures, no significant differences were found except for engagement in the Grammar Game, where children in the LbT condition reported significantly higher engagement than those in the Self-Practice condition ($p < 0.05$). 
    
    An exploratory analysis revealed a significant correlation between the time spent per iteration and perceived engagement in the LbT condition ($r=0.365, p<0.05$), whereas this effect was not observed in the Self-Practice condition ($r=-0.01, p>0.05$). Additionally, task difficulty was rated as moderate in both conditions, indicating that LbT was not perceived as more difficult than Self-Practice.

    \begin{table}[h]
        \centering
        \renewcommand{\arraystretch}{1.2} 
        \resizebox{\textwidth}{!}{%
            \begin{tabular}{lcccccc}
                \toprule
                \textbf{Variable} & \multicolumn{3}{c}{\textbf{Body Part Game}} & \multicolumn{3}{c}{\textbf{Grammar Game}} \\ 
                \cmidrule(lr){2-4} \cmidrule(lr){5-7}
                & \textbf{Learning by Teaching} & \textbf{Self-Practice} & \textbf{p-value} 
                & \textbf{Learning by Teaching} & \textbf{Self-Practice} & \textbf{p-value} \\ 
                \midrule
                Task Enjoyment       & 4.3 (0.83)  & 4.2 (0.81)  & 0.92  & 4.2 (0.97)  & 4.2 (1.07)  & 0.99  \\
                Perceived Competence & 4.0 (0.81)  & 4.3 (0.81)  & 0.41  & 4.3 (1.07)  & 4.4 (1.07)  & 0.95  \\
                Engagement           & 3.7 (1.07)  & 4.1 (0.99)  & 0.11  & \textbf{4.2 (0.99)}  & 3.3 (1.46)  & \textbf{0.03}  \\
                Difficulty           & 2.3 (0.94)  & 2.3 (1.06)  & 0.96  & 2.6 (1.26)  & 2.6 (1.46)  & 0.73  \\
                \bottomrule
            \end{tabular}
        }
        \caption{Comparison of Learning by Teaching and Self-Practice conditions across engagement, competence, enjoyment, and difficulty for both learning tasks. Significant differences are in bold.}
        \label{table:game_comparison}
    \end{table}

    \subsection{Influence of Prior Knowledge}
        We investigated whether students' prior knowledge influenced their learning gains in each condition. To do this, we split the data for each task and condition based on the median pre-test score. This resulted in two groups per task and condition: Low-Baseline (students with a pre-test score at or below the median) and High-Baseline (students scoring above the median).   The median scores for each group, along with demographic information (gender distribution and pre-test scores), are provided in the supplementary material

        Figure~\ref{fig:H3_post_gain} shows that Low-Baseline students achieved significantly higher post-gain compared to High-Baseline students in the Learning-by-Teaching condition, whereas this effect was not observed in the Self-Practice condition.

        Specifically, in the Body Parts Game, Low-Baseline students ($Mdn=4.00$) demonstrated significantly greater improvement after the intervention compared to High-Baseline students ($Mdn=0.00$), $U=52.00, z=-2.86, p<0.05, r=-0.51$. A similar pattern was observed in the Grammar Game, where Low-Baseline students ($Mdn=5.00$) showed significantly higher learning gains than High-Baseline students ($Mdn=0.00$), $U=29.00, z=-2.07, p<0.05, r=-0.44$.

        \begin{figure}[h]
        \centering
        \begin{subfigure}{0.48\textwidth}
            \centering
            \includegraphics[width=1\textwidth]{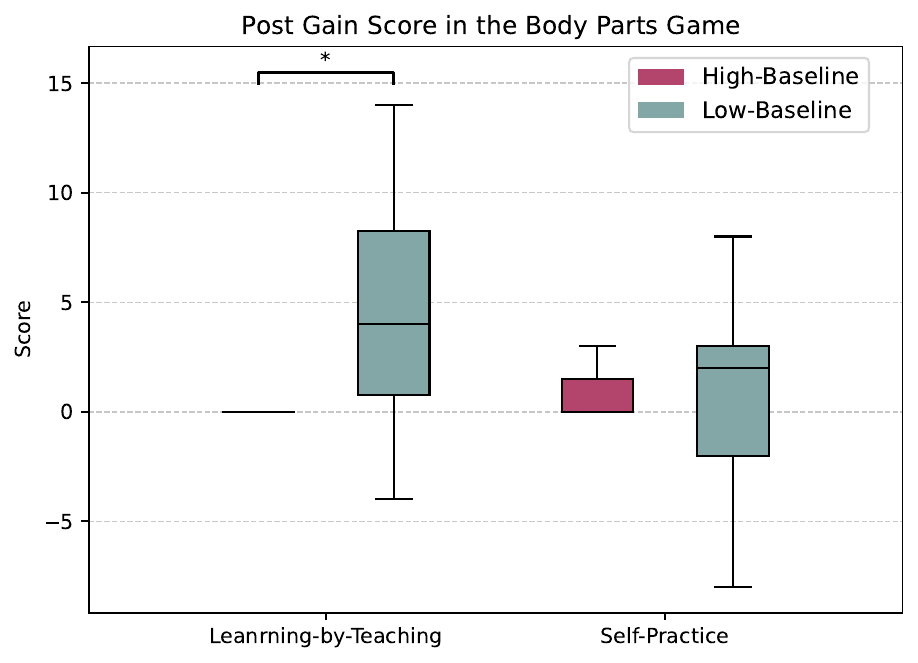}
            \caption{}
            \label{fig:H3_time_it}
        \end{subfigure}
        \begin{subfigure}{0.48\textwidth}
            \centering
            \includegraphics[width=1\textwidth]{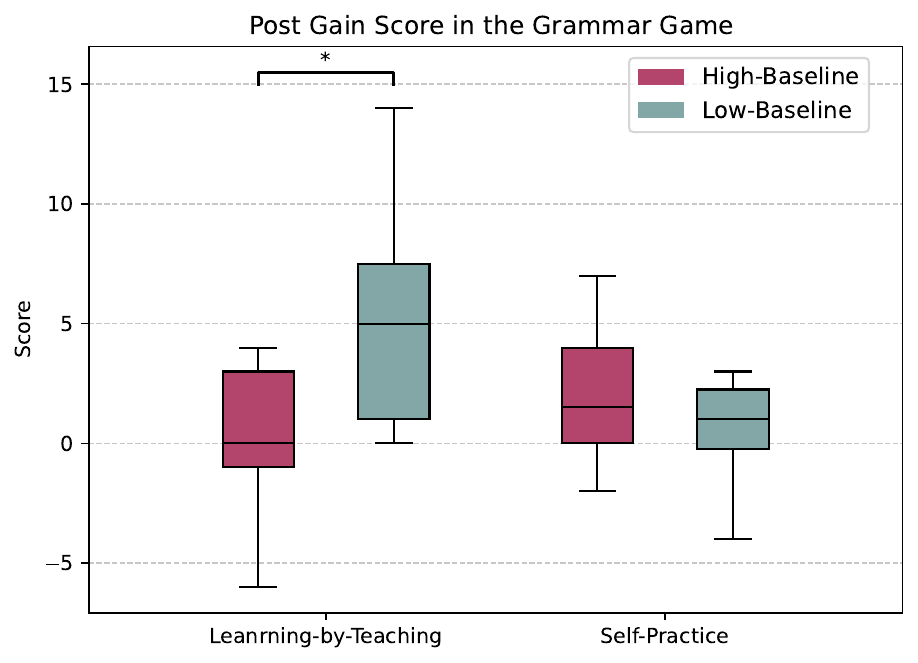}
            \caption{}
            \label{fig:H3_time_help}
        \end{subfigure}
        
        \caption{Post-test learning gains for Low-Baseline and High-Baseline students, comparing the Learning-by-Teaching and Self-Practice conditions. (a) Body Part Game. (b) Grammar Game. Low-Baseline students in the Learning-by-Teaching condition showed significantly greater improvements than their High-Baseline peers, while this effect was not observed in the Self-Practice condition.}
        \label{fig:H3_post_gain}
        \end{figure}
        
\begin{figure}[h]
        \centering
        \begin{subfigure}{0.48\textwidth}
            \centering
            \includegraphics[width=1\textwidth]{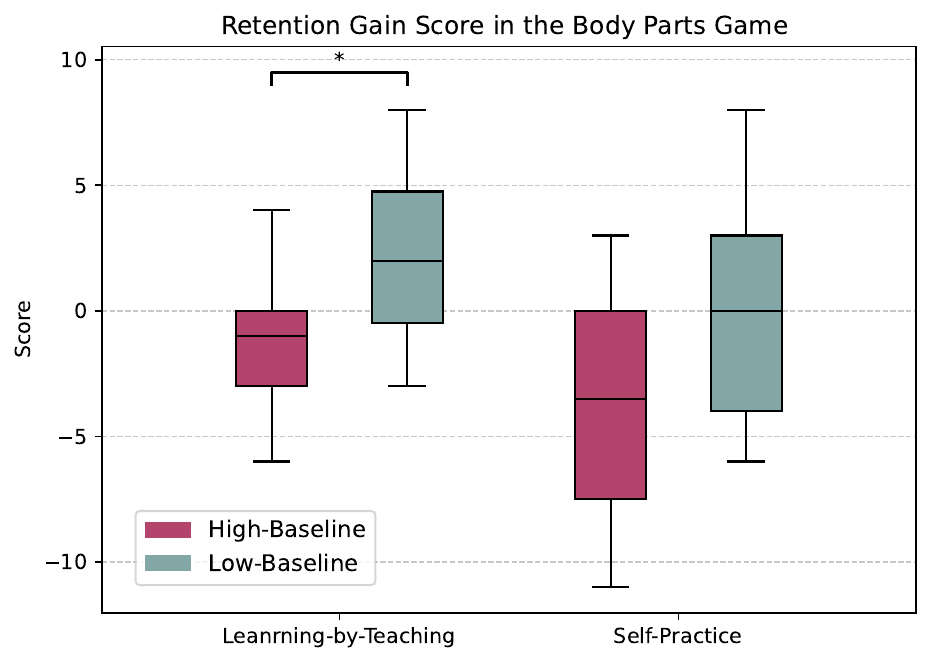}
            \caption{}
            
        \end{subfigure}
        \begin{subfigure}{0.48\textwidth}
            \centering
            \includegraphics[width=1\textwidth]{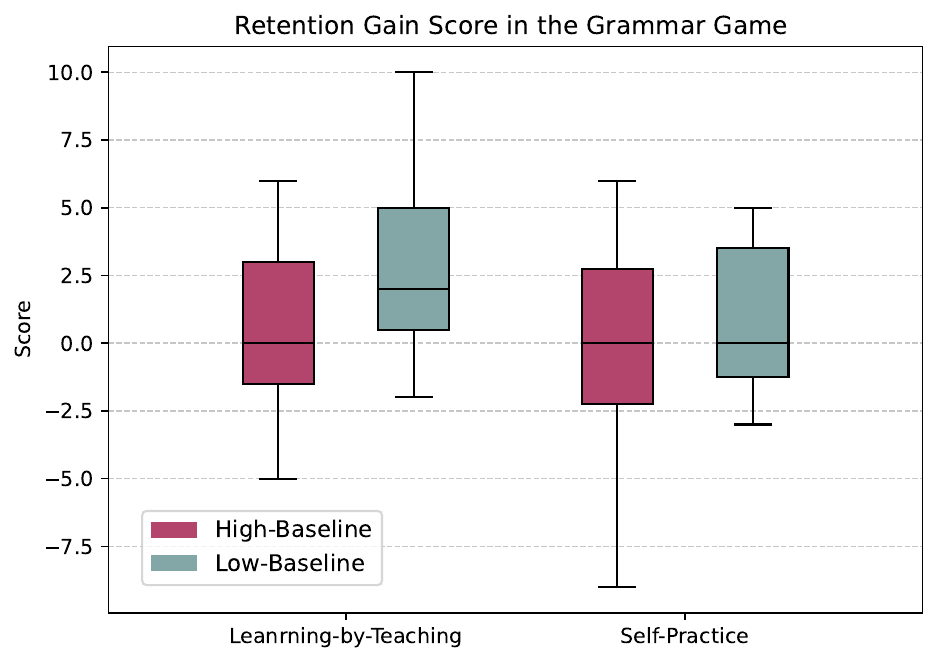}
            \caption{}
    
        \end{subfigure}
        
        \caption{Retention gain scores for the Learning-by-Teaching and Self-Practice conditions, separated by baseline knowledge groups. (a) Body Part Game: Low-Baseline students demonstrated significantly higher retention gains compared to High-Baseline students. (b) Grammar Game: While Low-Baseline students showed higher retention gains than their High-Baseline peers, the difference was not statistically significant but represented a medium effect size.}
        \label{fig:H3_retention_gain}
        \end{figure}

        Figure~\ref{fig:H3_retention_gain} shows that this effect also extended to Retention Gain, where Low-Baseline students demonstrated greater knowledge retention compared to their High-Baseline peers. This improvement was significant in the Body Parts Game, where Low-Baseline students ($Mdn=2.00$) retained more knowledge compared to High-Baseline students ($Mdn=-1.00$), $U=51.00, z=-3.37, r=-0.59$. A similar trend was observed in the Grammar Game, where Low-Baseline students ($Mdn=2.00$) outperformed High-Baseline students ($Mdn=0.00$) in retention gain. However, while this difference was not statistically significant ($U=40.00, p>0.05$), it did represent a medium effect size ($r=-0.44$).
        
        These results suggest that the Learning-by-Teaching condition was particularly beneficial for students with lower initial knowledge, enabling them to make greater learning gains compared to their higher-performing peers, which was not observed in the Self-Practice condition.

    \subsection{Influence of the nature of the task}
        We examined whether the nature of the task influenced children's interaction and perception of the learning activity in the Learning-by-Teaching condition. Regarding task enjoyment and perceived competence, no significant differences were found between the two tasks, with consistently high ratings across both, suggesting a potential ceiling effect. However, children in the Grammar Game ($Mdn = 4.67$) reported significantly higher engagement than those in the Body Part Game ($Mdn = 4.00$), $U=209.50, z=-2.08, r=-0.29$.  This suggest that tasks requiring rule inference, such as the Grammar Game, may promote deeper engagement and effortful learning, leading to stronger associations between time spent on the activity and perceived engagement.

        For interaction-related measures, no significant differences were found in the time spent per iteration or on the help panel between the two tasks. However, in the Grammar Game, time spent per iteration was strongly correlated with perceived task enjoyment ($r=0.56, p<0.01$), perceived competence ($r=0.41, p=0.05$), and perceived engagement ($r=0.46, p<0.05$), whereas no such correlations were observed in the Body Part Game. Additionally, the time spent on the help button in the Grammar Game was significantly correlated with retention gain ($r=0.61, p<0.01$), a relationship that was not observed in the Body Part Game ($r=0.02, p>0.05$). This could indicate the role of self-reflection and review in reinforcing learning outcomes in rule-inference games, a pattern not seen in the memorisation-based Body Part Game.

\section{Discussion}

    \subsection{Interactive RL as a Cognitive Model for Peer-Learning Robots}
        Our results indicate that children successfully assumed the tutor role by continuously providing accurate feedback to the robot, demonstrating their attentiveness and ability to evaluate its responses. These findings align with prior research showing that children can effectively guide a robot’s learning using evaluative feedback \citep{lemaignan2016learning, yadollahi2018deictic}.
        Furthermore, in the Body Part Game, children became more efficient tutors over time, as reflected in the progressive decrease in time spent per iteration. This adaptation aligns with research on LbT, where tutors refine their understanding through repeated explanations and assessments \citep{duran2017learning, roscoe2007understanding}. The reduction in help panel usage further suggests that children gradually relied less on external resources, reinforcing Interactive RL as an effective cognitive model for peer learning.
        However, in the Grammar Game, no significant reduction in help panel usage was observed. This suggests that grammar-related tasks require more complex cognitive processing, making it harder for children to rely solely on their prior knowledge \citep{biswas2005learning, roscoe2007understanding}.
        Overall, these findings validate H1, as children successfully assumed the tutor role, provided consistent and accurate feedback, and demonstrated adaptive learning strategies—particularly in the Body Part Game.

    \subsection{Learning-by-Teaching Enhances Engagement and Retention}
        Our study examined the impact of LbT a social robot compared to self-practice, focusing on both learning outcomes and engagement. While there was a trend toward higher learning gains in the Learning-by-Teaching condition, the difference was not statistically significant. One possible explanation is that children were informed in both the familiarisation and lesson phases that they would be teaching the robot. Prior research on LbT has demonstrated that even preparing to teach enhances learning outcomes, regardless of whether actual teaching occurs \citep{biswas2005learning, koh2018learning}. As a result, by the time children engaged in the intervention session, they may have already consolidated knowledge during the lesson phase, reducing the measurable difference between the Learning-by-Teaching and Self-Practice conditions. This observation suggests that the act of preparing to teach, even before engaging in the actual teaching process, may itself serve as a valuable cognitive activity, promoting initial consolidation and reducing the margin for additional gains during the intervention phase.
        
        Despite this, the retention gain in the Body Part Game was significantly higher in the LbT condition, suggesting that the social component of tutoring may support long-term knowledge retention. The social presence of the robot in the LbT condition may have contributed to higher engagement and cognitive effort, which aligns with prior research showing that embodied social agents enhance learning compared to virtual representations or traditional learning support \citep{leyzberg2012physical, kennedy2015robot, li2015benefit}. The physical presence of a robot, combined with its ability to use gestures and verbal interaction, fosters stronger social and cognitive engagement \citep{vrins2022you}.
        
        Additionally, social robots in educational settings have been shown to elicit caretaking behaviours and sustained attention \citep{tanaka2012children}. This social bonding effect may explain why children in our study spent significantly more time per iteration in the LbT condition compared to the Self-Practice condition. This aligns with the Protégé Effect , where learners invest more effort when they feel responsible for their tutee’s learning \citep{chase2009teachable}.
        Furthermore, children in the LbT condition spent more time using the help panel, particularly in the Grammar Game, where help panel usage was positively correlated with higher retention gain. This suggests that interacting with the robot encouraged deeper reflection and cognitive effort, reinforcing the idea that tutoring a social agent promotes active learning strategies \citep{roscoe2007understanding, biswas2005learning}.
        In terms of self-reported engagement, children rated the Grammar Game as more engaging in the LbT condition compared to the Self-Practice condition. However, no significant differences were found in task enjoyment or perceived competence. This is likely due to ceiling effects, a common challenge in studies with young children, where self-reported measures tend to cluster at the higher end of the scale, making it difficult to detect differences \citep{donnermann2022social}.
        Overall, our findings partially support H2, as children in the LbT condition showed higher engagement and better retention. However, we did not find sufficient evidence to conclude that this condition led to a more positive perception of the learning activity.

    \subsection{Children with lower prior knowledge benefit more from Learning-by-Teaching}
        Our results suggest that prior knowledge significantly influenced learning outcomes, particularly in the LbT condition. In both the Body Part Game and Grammar Game, Low-Baseline students (those with lower pre-test scores) demonstrated greater learning gains compared to High-Baseline students. This effect was statistically significant in the Body Part Game for both post-test and retention gain, while a similar trend was observed in the Grammar Game, though it did not reach significance.
        These findings align with previous research indicating that students with lower prior knowledge benefit more from LbT, as they engage more actively in self-explanation and error correction processes \citep{biswas2005learning, roscoe2007understanding}. Teaching a robot likely provided these students with structured opportunities to reflect on their understanding, helping them identify misconceptions and strengthen their grasp of the material. 
        Interestingly, the Self-Practice condition did not yield a significant difference between Low- and High-Baseline students. This suggests that students with lower prior knowledge may not have engaged as deeply with the material when practicing independently. In contrast, the social and interactive nature of the LbT condition may have provided scaffolding that supported deeper learning, particularly for those with weaker initial knowledge.
        However, it is important to consider that High-Baseline students had less room for improvement, as their initial knowledge levels were already relatively high. This ceiling effect may partially explain the lower learning gains observed in this group. 
        These findings partially support H3, showing that LbT is particularly beneficial for students with lower prior knowledge, enabling them to achieve greater learning gains compared to self-practice.

    \subsection{Rule-Inference Tasks Are More Effective for Learning-by-Teaching}
        Our results indicate that the nature of the learning task influenced children's interaction and perception of the activity in the LbT condition. Specifically, children in the Grammar Game reported significantly higher engagement compared to those in the Body Part Game, suggesting that the cognitive demands of the task may have led to greater immersion in the learning process. This aligns with research suggesting that effortful learning can enhance engagement and satisfaction when learners perceive progress \citep{bjork2011making}.
        However, no significant differences were found in task enjoyment or perceived competence between the two tasks. This is likely due to ceiling effects as explained in the previous section.  
        Regarding interaction patterns, we found no significant difference in time spent per iteration or on the help panel between the two tasks. However, correlational analyses revealed notable differences in how children engaged with the learning activity. In the Grammar Game, time spent per iteration was positively correlated with task enjoyment, perceived competence, and engagement, while no such relationships were observed in the Body Part Game. Additionally, time spent on the help panel in the Grammar Game was strongly correlated with retention gain, suggesting that children who spent more time reviewing material during the activity retained knowledge more effectively. This was not observed in the Body Part Game.
        These findings suggest that the Grammar Game elicited deeper cognitive engagement, possibly because it required rule-based reasoning rather than simple memorisation. Prior research has highlighted that LbT is particularly effective for tasks that involve higher-order thinking, such as conceptual reasoning and problem-solving \citep{biswas2005learning, roscoe2007understanding}. The reduced time on the help panel usage for the Body Part Game further supports the idea that this task relied more on rote learning, reducing the need for active review during the game.
        Overall, our results partially support H4, demonstrating that the Grammar Game elicited higher engagement and deeper cognitive processing, particularly in relation to time spent on the help panel and retention gain. However, the ceiling effects in self-reported measures of enjoyment and competence prevented us from fully validating the hypothesis.

\section{Conclusion, Limitation \& Future Work}
    \paragraph{Limitations} While our study provides valuable insights into the effectiveness of Learning-by-Teaching a social robot, several limitations should be acknowledged. First, study was conducted in a single primary school, which may limit the generalisability of the findings to other educational contexts or diverse student populations. Future studies should aim to replicate the experiment in multiple schools, including those with different socio-economic backgrounds, to assess the robustness and generalisability of the results. Moreover, self-reported measures of engagement, enjoyment, and competence exhibited ceiling effects, which is common in studies involving young children \citep{donnermann2022social}. This may have masked potential differences between conditions. To address this, future work should incorporate alternative assessment methods, such as behavioural engagement metrics (e.g., gaze tracking, facial expressions, or speech analysis), to provide a more nuanced understanding of children's experiences \citep{aitsam2024measuring}. 

    \paragraph{Ethical Considerations and Well-Being} While educational technologies offer novel ways to engage learners, recent research highlights concerns about their potential negative effects on children’s physical, cognitive, emotional, and social well-being \citep{melo2020educational}. Excessive screen time, overstimulation, and addictive design elements in gamified learning environments have been associated with sleep problems, attention difficulties, and increased stress. In response to this growing concern, scholars have called for more responsible integration of digital tools in classrooms, with an emphasis on sustainable use and long-term learner welfare.
    Our study contributes to this conversation by exploring an alternative approach: the use of a physically embodied social robot designed to emulate peer-like behaviour. Unlike traditional screen-based platforms, social robots provide a form of embodied interaction that can encourage more balanced, socially grounded experiences. The robot’s physical presence, gesture use, and adaptive responsiveness may help counteract some of the isolation or overstimulation associated with purely screen-based systems. Furthermore, the LbT paradigm promotes child agency, responsibility, and reflection; all of which are linked to healthy cognitive and emotional development.
    Nevertheless, we recognise that the inclusion of robot peers does not automatically resolve all concerns. As gamification and interaction design increasingly shape educational experiences, future research should systematically investigate how robot-mediated learning affects children’s well-being over time. This includes assessing not only learning outcomes, but also emotional resilience, social connection, and behavioural regulation, particularly in relation to screen-based alternatives.

    \paragraph{Conclusion} This study introduces Interactive RL as a cognitive model for peer-learning robots in real classroom environments. By enabling the robot to update its behaviour in response to children's evaluative feedback, our approach supports a bi-directional, learning-by-teaching dynamic that goes beyond scripted or pre-trained robot responses. Children were able to effectively tutor the robot, offering accurate and consistent feedback while adapting their strategies over time; supporting the feasibility of Interactive RL-powered peer robots in classroom contexts. Compared to traditional self-practice, the learning-by-teaching condition led to significantly higher retention gains, particularly in the Grammar Game. This suggests that tutoring a robot promotes deeper processing and long-term consolidation of knowledge.
    Our analysis also showed that children with lower prior knowledge benefited the most from teaching the robot, highlighting the potential of teachable robots to act as personalised scaffolding tools that support equity in learning. Furthermore, the task type influenced outcomes: rule-inference tasks like the Grammar Game fostered greater engagement and stronger retention than memorisation-based tasks, reinforcing the role of cognitive depth in LbT scenarios. This work also demonstrates the practical feasibility of deploying multiple autonomous robots simultaneously in classrooms, offering both ecological validity and insights into scalable implementation. These findings suggest that Interactive RL can transform social robots from passive learners into active, adaptive partners, opening promising directions for future AI-powered, collaborative learning environments.

\section*{Data Availability}
The data supporting the findings of this study are available from the corresponding author upon reasonable request.

\bibliographystyle{elsarticle-harv} 
\bibliography{ref}

\begin{thebibliography}{50}
\expandafter\ifx\csname natexlab\endcsname\relax\def\natexlab#1{#1}\fi
\providecommand{\url}[1]{\texttt{#1}}
\providecommand{\href}[2]{#2}
\providecommand{\path}[1]{#1}
\providecommand{\DOIprefix}{doi:}
\providecommand{\ArXivprefix}{arXiv:}
\providecommand{\URLprefix}{URL: }
\providecommand{\Pubmedprefix}{pmid:}
\providecommand{\doi}[1]{\href{http://dx.doi.org/#1}{\path{#1}}}
\providecommand{\Pubmed}[1]{\href{pmid:#1}{\path{#1}}}
\providecommand{\bibinfo}[2]{#2}
\ifx\xfnm\relax \def\xfnm[#1]{\unskip,\space#1}\fi
\bibitem[{Aitsam et~al.(2024)Aitsam, Lacroix, Goyal, Bartolozzi and Di~Nuovo}]{aitsam2024measuring}
\bibinfo{author}{Aitsam, M.}, \bibinfo{author}{Lacroix, D.}, \bibinfo{author}{Goyal, G.}, \bibinfo{author}{Bartolozzi, C.}, \bibinfo{author}{Di~Nuovo, A.}, \bibinfo{year}{2024}.
\newblock \bibinfo{title}{Measuring cognitive load through event camera based human-pose estimation}, in: \bibinfo{booktitle}{International Workshop on Human-Friendly Robotics}, \bibinfo{organization}{Springer}. pp. \bibinfo{pages}{229--239}.
\bibitem[{Alimardani et~al.(2022)Alimardani, Harinandansingh, Ravin and De~Haas}]{alimardani2022motivational}
\bibinfo{author}{Alimardani, M.}, \bibinfo{author}{Harinandansingh, J.}, \bibinfo{author}{Ravin, L.}, \bibinfo{author}{De~Haas, M.}, \bibinfo{year}{2022}.
\newblock \bibinfo{title}{Motivational gestures in robot-assisted language learning: a study of cognitive engagement using eeg brain activity}, in: \bibinfo{booktitle}{2022 31st IEEE International Conference on Robot and Human Interactive Communication (RO-MAN)}, \bibinfo{organization}{IEEE}. pp. \bibinfo{pages}{1393--1398}.
\bibitem[{Almousa and Alghowinem(2023)}]{almousa2023conceptualization}
\bibinfo{author}{Almousa, O.}, \bibinfo{author}{Alghowinem, S.}, \bibinfo{year}{2023}.
\newblock \bibinfo{title}{Conceptualization and development of an autonomous and personalized early literacy content and robot tutor behavior for preschool children}.
\newblock \bibinfo{journal}{User Modeling and User-Adapted Interaction} \bibinfo{volume}{33}, \bibinfo{pages}{261--291}.
\bibitem[{Authors(2023)}]{Authors2023}
\bibinfo{author}{Authors}, \bibinfo{year}{2023}.
\newblock \bibinfo{title}{To be added following double-blind review.}
\bibitem[{Belpaeme et~al.(2018)Belpaeme, Kennedy, Ramachandran, Scassellati and Tanaka}]{belpaeme2018social}
\bibinfo{author}{Belpaeme, T.}, \bibinfo{author}{Kennedy, J.}, \bibinfo{author}{Ramachandran, A.}, \bibinfo{author}{Scassellati, B.}, \bibinfo{author}{Tanaka, F.}, \bibinfo{year}{2018}.
\newblock \bibinfo{title}{Social robots for education: A review}.
\newblock \bibinfo{journal}{Science robotics} \bibinfo{volume}{3}, \bibinfo{pages}{eaat5954}.
\bibitem[{Biswas et~al.(2005)Biswas, Leelawong, Schwartz, Vye and at~Vanderbilt}]{biswas2005learning}
\bibinfo{author}{Biswas, G.}, \bibinfo{author}{Leelawong, K.}, \bibinfo{author}{Schwartz, D.}, \bibinfo{author}{Vye, N.}, \bibinfo{author}{at~Vanderbilt, T.T.A.G.}, \bibinfo{year}{2005}.
\newblock \bibinfo{title}{Learning by teaching: A new agent paradigm for educational software}.
\newblock \bibinfo{journal}{Applied Artificial Intelligence} \bibinfo{volume}{19}, \bibinfo{pages}{363--392}.
\bibitem[{Bjork et~al.(2011)Bjork, Bjork et~al.}]{bjork2011making}
\bibinfo{author}{Bjork, E.L.}, \bibinfo{author}{Bjork, R.A.}, et~al., \bibinfo{year}{2011}.
\newblock \bibinfo{title}{Making things hard on yourself, but in a good way: Creating desirable difficulties to enhance learning}.
\newblock \bibinfo{journal}{Psychology and the real world: Essays illustrating fundamental contributions to society} \bibinfo{volume}{2}.
\bibitem[{Bruzzo et~al.(2024)Bruzzo, Matarese, Sciutti and Rea}]{bruzzo2024charm}
\bibinfo{author}{Bruzzo, D.}, \bibinfo{author}{Matarese, M.}, \bibinfo{author}{Sciutti, A.}, \bibinfo{author}{Rea, F.}, \bibinfo{year}{2024}.
\newblock \bibinfo{title}{Charm or harm? how social robotic tutors influence people’s learning with correct and incorrect guidance}, in: \bibinfo{booktitle}{International Conference on Social Robotics}, \bibinfo{organization}{Springer}. pp. \bibinfo{pages}{475--487}.
\bibitem[{Chandra et~al.(2020)Chandra, Dillenbourg and Paiva}]{chandra2020children}
\bibinfo{author}{Chandra, S.}, \bibinfo{author}{Dillenbourg, P.}, \bibinfo{author}{Paiva, A.}, \bibinfo{year}{2020}.
\newblock \bibinfo{title}{Children teach handwriting to a social robot with different learning competencies}.
\newblock \bibinfo{journal}{International Journal of Social Robotics} \bibinfo{volume}{12}, \bibinfo{pages}{721--748}.
\bibitem[{Chase et~al.(2009)Chase, Chin, Oppezzo and Schwartz}]{chase2009teachable}
\bibinfo{author}{Chase, C.C.}, \bibinfo{author}{Chin, D.B.}, \bibinfo{author}{Oppezzo, M.A.}, \bibinfo{author}{Schwartz, D.L.}, \bibinfo{year}{2009}.
\newblock \bibinfo{title}{Teachable agents and the prot{\'e}g{\'e} effect: Increasing the effort towards learning}.
\newblock \bibinfo{journal}{Journal of science education and technology} \bibinfo{volume}{18}, \bibinfo{pages}{334--352}.
\bibitem[{Chen et~al.(2020)Chen, Park and Breazeal}]{chen2020teaching}
\bibinfo{author}{Chen, H.}, \bibinfo{author}{Park, H.W.}, \bibinfo{author}{Breazeal, C.}, \bibinfo{year}{2020}.
\newblock \bibinfo{title}{Teaching and learning with children: Impact of reciprocal peer learning with a social robot on children’s learning and emotive engagement}.
\newblock \bibinfo{journal}{Computers \& Education} \bibinfo{volume}{150}, \bibinfo{pages}{103836}.
\bibitem[{Chen and Liu(2024)}]{chen2024impact}
\bibinfo{author}{Chen, P.Y.}, \bibinfo{author}{Liu, Y.C.}, \bibinfo{year}{2024}.
\newblock \bibinfo{title}{Impact of ai robot image recognition technology on improving students' conceptual understanding of cell division and science learning motivation.}
\newblock \bibinfo{journal}{Journal of Baltic Science Education} \bibinfo{volume}{23}, \bibinfo{pages}{208--220}.
\bibitem[{Chiang et~al.(2023)Chiang, Cheng and Chen}]{chiang2023improving}
\bibinfo{author}{Chiang, Y.h.V.}, \bibinfo{author}{Cheng, Y.W.}, \bibinfo{author}{Chen, N.S.}, \bibinfo{year}{2023}.
\newblock \bibinfo{title}{Improving language learning activity design through identifying learning difficulties in a platform using educational robots and iot-based tangible objects}.
\newblock \bibinfo{journal}{Educational Technology \& Society} \bibinfo{volume}{26}, \bibinfo{pages}{84--100}.
\bibitem[{Das and Pon-Barry(2018)}]{das-pon-barry-2018-turn}
\bibinfo{author}{Das, R.}, \bibinfo{author}{Pon-Barry, H.}, \bibinfo{year}{2018}.
\newblock \bibinfo{title}{Turn-taking strategies for human-robot peer-learning dialogue}, in: \bibinfo{editor}{Komatani, K.}, \bibinfo{editor}{Litman, D.}, \bibinfo{editor}{Yu, K.}, \bibinfo{editor}{Papangelis, A.}, \bibinfo{editor}{Cavedon, L.}, \bibinfo{editor}{Nakano, M.} (Eds.), \bibinfo{booktitle}{Proceedings of the 19th Annual {SIG}dial Meeting on Discourse and Dialogue}, \bibinfo{publisher}{Association for Computational Linguistics}, \bibinfo{address}{Melbourne, Australia}. pp. \bibinfo{pages}{119--129}.
\newblock \URLprefix \url{https://aclanthology.org/W18-5013/}, \DOIprefix\doi{10.18653/v1/W18-5013}.
\bibitem[{Demir-Lira et~al.(2020)Demir-Lira, Kanero, Oran{\c{c}}, Ko{\c{s}}kulu, Franko, G{\"o}ksun and K{\"u}ntay}]{demir2020l2}
\bibinfo{author}{Demir-Lira, {\"O}.E.}, \bibinfo{author}{Kanero, J.}, \bibinfo{author}{Oran{\c{c}}, C.}, \bibinfo{author}{Ko{\c{s}}kulu, S.}, \bibinfo{author}{Franko, I.}, \bibinfo{author}{G{\"o}ksun, T.}, \bibinfo{author}{K{\"u}ntay, A.C.}, \bibinfo{year}{2020}.
\newblock \bibinfo{title}{L2 vocabulary teaching by social robots: The role of gestures and on-screen cues as scaffolds}, in: \bibinfo{booktitle}{Frontiers in education}, \bibinfo{organization}{Frontiers Media SA}. p. \bibinfo{pages}{599636}.
\bibitem[{Donnermann and Lugrin(2024)}]{donnermann2024integration}
\bibinfo{author}{Donnermann, M.}, \bibinfo{author}{Lugrin, B.}, \bibinfo{year}{2024}.
\newblock \bibinfo{title}{Integration of robot-supported tutoring in higher education-an empirically based concept}, in: \bibinfo{booktitle}{Proceedings of the 2024 the 16th International Conference on Education Technology and Computers}, pp. \bibinfo{pages}{1--7}.
\bibitem[{Donnermann et~al.(2022)Donnermann, Schaper and Lugrin}]{donnermann2022social}
\bibinfo{author}{Donnermann, M.}, \bibinfo{author}{Schaper, P.}, \bibinfo{author}{Lugrin, B.}, \bibinfo{year}{2022}.
\newblock \bibinfo{title}{Social robots in applied settings: A long-term study on adaptive robotic tutors in higher education}.
\newblock \bibinfo{journal}{Frontiers in Robotics and AI} \bibinfo{volume}{9}, \bibinfo{pages}{831633}.
\bibitem[{Duran(2017)}]{duran2017learning}
\bibinfo{author}{Duran, D.}, \bibinfo{year}{2017}.
\newblock \bibinfo{title}{Learning-by-teaching. evidence and implications as a pedagogical mechanism}.
\newblock \bibinfo{journal}{Innovations in education and teaching international} \bibinfo{volume}{54}, \bibinfo{pages}{476--484}.
\bibitem[{El~Hamamsy et~al.(2019)El~Hamamsy, Johal, Asselborn, Nasir and Dillenbourg}]{el2019learning}
\bibinfo{author}{El~Hamamsy, L.}, \bibinfo{author}{Johal, W.}, \bibinfo{author}{Asselborn, T.}, \bibinfo{author}{Nasir, J.}, \bibinfo{author}{Dillenbourg, P.}, \bibinfo{year}{2019}.
\newblock \bibinfo{title}{Learning by collaborative teaching: an engaging multi-party cowriter activity}, in: \bibinfo{booktitle}{2019 28th IEEE international conference on robot and human interactive communication (RO-MAN)}, \bibinfo{organization}{IEEE}. pp. \bibinfo{pages}{1--8}.
\bibitem[{Henkel et~al.(2025)Henkel, Horne-Robinson, Hills, Roberts and McGrane}]{henkel2025supporting}
\bibinfo{author}{Henkel, O.}, \bibinfo{author}{Horne-Robinson, H.}, \bibinfo{author}{Hills, L.}, \bibinfo{author}{Roberts, B.}, \bibinfo{author}{McGrane, J.}, \bibinfo{year}{2025}.
\newblock \bibinfo{title}{Supporting literacy assessment in west africa: Using state-of-the-art speech models to assess oral reading fluency}.
\newblock \bibinfo{journal}{International Journal of Artificial Intelligence in Education} , \bibinfo{pages}{1--22}.
\bibitem[{Kennedy et~al.(2015)Kennedy, Baxter and Belpaeme}]{kennedy2015robot}
\bibinfo{author}{Kennedy, J.}, \bibinfo{author}{Baxter, P.}, \bibinfo{author}{Belpaeme, T.}, \bibinfo{year}{2015}.
\newblock \bibinfo{title}{The robot who tried too hard: Social behaviour of a robot tutor can negatively affect child learning}, in: \bibinfo{booktitle}{Proceedings of the tenth annual ACM/IEEE international conference on human-robot interaction}, pp. \bibinfo{pages}{67--74}.
\bibitem[{Knox and Stone(2009)}]{knox2009interactively}
\bibinfo{author}{Knox, W.B.}, \bibinfo{author}{Stone, P.}, \bibinfo{year}{2009}.
\newblock \bibinfo{title}{Interactively shaping agents via human reinforcement: The tamer framework}, in: \bibinfo{booktitle}{Proceedings of the fifth international conference on Knowledge capture}, pp. \bibinfo{pages}{9--16}.
\bibitem[{Koh et~al.(2018)Koh, Lee and Lim}]{koh2018learning}
\bibinfo{author}{Koh, A.W.L.}, \bibinfo{author}{Lee, S.C.}, \bibinfo{author}{Lim, S.W.H.}, \bibinfo{year}{2018}.
\newblock \bibinfo{title}{The learning benefits of teaching: A retrieval practice hypothesis}.
\newblock \bibinfo{journal}{Applied Cognitive Psychology} \bibinfo{volume}{32}, \bibinfo{pages}{401--410}.
\bibitem[{Lee et~al.(2021)Lee, Chauhan, Goh, Nilsen and Law}]{lee2021curiosity}
\bibinfo{author}{Lee, K.J.}, \bibinfo{author}{Chauhan, A.}, \bibinfo{author}{Goh, J.}, \bibinfo{author}{Nilsen, E.}, \bibinfo{author}{Law, E.}, \bibinfo{year}{2021}.
\newblock \bibinfo{title}{Curiosity notebook: the design of a research platform for learning by teaching}.
\newblock \bibinfo{journal}{Proceedings of the ACM on Human-Computer Interaction} \bibinfo{volume}{5}, \bibinfo{pages}{1--26}.
\bibitem[{Lemaignan et~al.(2016)Lemaignan, Jacq, Hood, Garcia, Paiva and Dillenbourg}]{lemaignan2016learning}
\bibinfo{author}{Lemaignan, S.}, \bibinfo{author}{Jacq, A.}, \bibinfo{author}{Hood, D.}, \bibinfo{author}{Garcia, F.}, \bibinfo{author}{Paiva, A.}, \bibinfo{author}{Dillenbourg, P.}, \bibinfo{year}{2016}.
\newblock \bibinfo{title}{Learning by teaching a robot: The case of handwriting}.
\newblock \bibinfo{journal}{IEEE Robotics \& Automation Magazine} \bibinfo{volume}{23}, \bibinfo{pages}{56--66}.
\bibitem[{Leyzberg et~al.(2012)Leyzberg, Spaulding, Toneva and Scassellati}]{leyzberg2012physical}
\bibinfo{author}{Leyzberg, D.}, \bibinfo{author}{Spaulding, S.}, \bibinfo{author}{Toneva, M.}, \bibinfo{author}{Scassellati, B.}, \bibinfo{year}{2012}.
\newblock \bibinfo{title}{The physical presence of a robot tutor increases cognitive learning gains}, in: \bibinfo{booktitle}{Proceedings of the annual meeting of the cognitive science society}.
\bibitem[{Li(2015)}]{li2015benefit}
\bibinfo{author}{Li, J.}, \bibinfo{year}{2015}.
\newblock \bibinfo{title}{The benefit of being physically present: A survey of experimental works comparing copresent robots, telepresent robots and virtual agents}.
\newblock \bibinfo{journal}{International Journal of Human-Computer Studies} \bibinfo{volume}{77}, \bibinfo{pages}{23--37}.
\bibitem[{Magpusao(2024)}]{magpusao2024gamification}
\bibinfo{author}{Magpusao, J.}, \bibinfo{year}{2024}.
\newblock \bibinfo{title}{Gamification and game-based learning in primary education: A bibliometric analysis}.
\newblock \bibinfo{journal}{Computers and Children} \bibinfo{volume}{3}.
\bibitem[{Melo et~al.(2020)Melo, Madariaga, Nussbaum, Heller, Bennett, Tsai and van Braak}]{melo2020educational}
\bibinfo{author}{Melo, C.}, \bibinfo{author}{Madariaga, L.}, \bibinfo{author}{Nussbaum, M.}, \bibinfo{author}{Heller, R.}, \bibinfo{author}{Bennett, S.}, \bibinfo{author}{Tsai, C.C.}, \bibinfo{author}{van Braak, J.}, \bibinfo{year}{2020}.
\newblock \bibinfo{title}{Educational technology and addictions}.
\bibitem[{Najar and Chetouani(2021)}]{najar2021reinforcement}
\bibinfo{author}{Najar, A.}, \bibinfo{author}{Chetouani, M.}, \bibinfo{year}{2021}.
\newblock \bibinfo{title}{Reinforcement learning with human advice: a survey}.
\newblock \bibinfo{journal}{Frontiers in Robotics and AI} \bibinfo{volume}{8}, \bibinfo{pages}{584075}.
\bibitem[{Nasir et~al.(2024)Nasir, Bruno and Dillenbourg}]{nasir2024social}
\bibinfo{author}{Nasir, J.}, \bibinfo{author}{Bruno, B.}, \bibinfo{author}{Dillenbourg, P.}, \bibinfo{year}{2024}.
\newblock \bibinfo{title}{Social robots as skilled ignorant peers for supporting learning}.
\newblock \bibinfo{journal}{Frontiers in Robotics and AI} \bibinfo{volume}{11}, \bibinfo{pages}{1385780}.
\bibitem[{Obayashi et~al.(2000)Obayashi, Shimoda and Yoshikawa}]{obayashi2000construction}
\bibinfo{author}{Obayashi, F.}, \bibinfo{author}{Shimoda, H.}, \bibinfo{author}{Yoshikawa, H.}, \bibinfo{year}{2000}.
\newblock \bibinfo{title}{Construction and evaluation of cai system based on learning by teaching to virtual student}, in: \bibinfo{booktitle}{Proceedings of the World Multiconference on Systemics, Cybernetics and Informatics}, pp. \bibinfo{pages}{94--99}.
\bibitem[{Okazaki et~al.(2015)Okazaki, Kanai, Ogata, Hasegawa, Ishii and Imai}]{okazaki2015building}
\bibinfo{author}{Okazaki, H.}, \bibinfo{author}{Kanai, Y.}, \bibinfo{author}{Ogata, M.}, \bibinfo{author}{Hasegawa, K.}, \bibinfo{author}{Ishii, K.}, \bibinfo{author}{Imai, M.}, \bibinfo{year}{2015}.
\newblock \bibinfo{title}{Building pedagogical relationships between humans and robots in natural interactions}, in: \bibinfo{booktitle}{Proceedings of the 3rd International Conference on Human-Agent Interaction}, pp. \bibinfo{pages}{115--120}.
\bibitem[{Pareto et~al.(2022)Pareto, Ekstr{\"o}m and Serholt}]{pareto2022children}
\bibinfo{author}{Pareto, L.}, \bibinfo{author}{Ekstr{\"o}m, S.}, \bibinfo{author}{Serholt, S.}, \bibinfo{year}{2022}.
\newblock \bibinfo{title}{Children’s learning-by-teaching with a social robot versus a younger child: Comparing interactions and tutoring styles}.
\newblock \bibinfo{journal}{Frontiers in Robotics and AI} \bibinfo{volume}{9}, \bibinfo{pages}{875704}.
\bibitem[{Roscoe and Chi(2007)}]{roscoe2007understanding}
\bibinfo{author}{Roscoe, R.D.}, \bibinfo{author}{Chi, M.T.}, \bibinfo{year}{2007}.
\newblock \bibinfo{title}{Understanding tutor learning: Knowledge-building and knowledge-telling in peer tutors’ explanations and questions}.
\newblock \bibinfo{journal}{Review of educational research} \bibinfo{volume}{77}, \bibinfo{pages}{534--574}.
\bibitem[{Roscoe and Chi(2008)}]{roscoe2008tutor}
\bibinfo{author}{Roscoe, R.D.}, \bibinfo{author}{Chi, M.T.}, \bibinfo{year}{2008}.
\newblock \bibinfo{title}{Tutor learning: The role of explaining and responding to questions}.
\newblock \bibinfo{journal}{Instructional science} \bibinfo{volume}{36}, \bibinfo{pages}{321--350}.
\bibitem[{Ryan et~al.(1983)Ryan, Mims and Koestner}]{ryan1983relation}
\bibinfo{author}{Ryan, R.M.}, \bibinfo{author}{Mims, V.}, \bibinfo{author}{Koestner, R.}, \bibinfo{year}{1983}.
\newblock \bibinfo{title}{Relation of reward contingency and interpersonal context to intrinsic motivation: A review and test using cognitive evaluation theory.}
\newblock \bibinfo{journal}{Journal of personality and Social Psychology} \bibinfo{volume}{45}, \bibinfo{pages}{736}.
\bibitem[{Serholt et~al.(2022)Serholt, Ekstr{\"o}m, K{\"u}ster, Ljungblad and Pareto}]{serholt2022comparing}
\bibinfo{author}{Serholt, S.}, \bibinfo{author}{Ekstr{\"o}m, S.}, \bibinfo{author}{K{\"u}ster, D.}, \bibinfo{author}{Ljungblad, S.}, \bibinfo{author}{Pareto, L.}, \bibinfo{year}{2022}.
\newblock \bibinfo{title}{Comparing a robot tutee to a human tutee in a learning-by-teaching scenario with children}.
\newblock \bibinfo{journal}{Frontiers in Robotics and AI} \bibinfo{volume}{9}, \bibinfo{pages}{836462}.
\bibitem[{Song et~al.(2024)Song, Barakova, Ham and Markopoulos}]{song2024impact}
\bibinfo{author}{Song, H.}, \bibinfo{author}{Barakova, E.I.}, \bibinfo{author}{Ham, J.}, \bibinfo{author}{Markopoulos, P.}, \bibinfo{year}{2024}.
\newblock \bibinfo{title}{The impact of social robots' presence and roles on children's performance in musical instrument practice}.
\newblock \bibinfo{journal}{British Journal of Educational Technology} \bibinfo{volume}{55}, \bibinfo{pages}{1041--1059}.
\bibitem[{Tanaka et~al.(2007)Tanaka, Cicourel and Movellan}]{tanaka2007socialization}
\bibinfo{author}{Tanaka, F.}, \bibinfo{author}{Cicourel, A.}, \bibinfo{author}{Movellan, J.R.}, \bibinfo{year}{2007}.
\newblock \bibinfo{title}{Socialization between toddlers and robots at an early childhood education center}.
\newblock \bibinfo{journal}{Proceedings of the National Academy of Sciences} \bibinfo{volume}{104}, \bibinfo{pages}{17954--17958}.
\bibitem[{Tanaka and Matsuzoe(2012)}]{tanaka2012children}
\bibinfo{author}{Tanaka, F.}, \bibinfo{author}{Matsuzoe, S.}, \bibinfo{year}{2012}.
\newblock \bibinfo{title}{Children teach a care-receiving robot to promote their learning: Field experiments in a classroom for vocabulary learning}.
\newblock \bibinfo{journal}{Journal of Human-Robot Interaction} \bibinfo{volume}{1}, \bibinfo{pages}{78--95}.
\bibitem[{Tu et~al.(2025)Tu, Chen and Huang}]{tu2025empowering}
\bibinfo{author}{Tu, Y.}, \bibinfo{author}{Chen, J.}, \bibinfo{author}{Huang, C.}, \bibinfo{year}{2025}.
\newblock \bibinfo{title}{Empowering personalized learning with generative artificial intelligence: Mechanisms, challenges and pathways}.
\newblock \bibinfo{journal}{Frontiers of Digital Education} \bibinfo{volume}{2}, \bibinfo{pages}{1--18}.
\bibitem[{Verhoeven et~al.(2018)Verhoeven, Catala and Theune}]{verhoeven2018designing}
\bibinfo{author}{Verhoeven, G.}, \bibinfo{author}{Catala, A.}, \bibinfo{author}{Theune, M.}, \bibinfo{year}{2018}.
\newblock \bibinfo{title}{Designing a playful robot application for second language learning}, in: \bibinfo{booktitle}{International Conference on ArtsIT, Interactivity and Game Creation}, \bibinfo{organization}{Springer}. pp. \bibinfo{pages}{385--394}.
\bibitem[{Verhoeven et~al.(2019)Verhoeven, Catala and Theune}]{10.1007/978-3-030-06134-0_42}
\bibinfo{author}{Verhoeven, G.}, \bibinfo{author}{Catala, A.}, \bibinfo{author}{Theune, M.}, \bibinfo{year}{2019}.
\newblock \bibinfo{title}{Designing a playful robot application for second language learning}, in: \bibinfo{editor}{Brooks, A.L.}, \bibinfo{editor}{Brooks, E.}, \bibinfo{editor}{Sylla, C.} (Eds.), \bibinfo{booktitle}{Interactivity, Game Creation, Design, Learning, and Innovation}, \bibinfo{publisher}{Springer International Publishing}, \bibinfo{address}{Cham}. pp. \bibinfo{pages}{385--394}.
\bibitem[{Vrins et~al.(2022)Vrins, Pruss, Prinsen, Ceccato and Alimardani}]{vrins2022you}
\bibinfo{author}{Vrins, A.}, \bibinfo{author}{Pruss, E.}, \bibinfo{author}{Prinsen, J.}, \bibinfo{author}{Ceccato, C.}, \bibinfo{author}{Alimardani, M.}, \bibinfo{year}{2022}.
\newblock \bibinfo{title}{Are you paying attention? the effect of embodied interaction with an adaptive robot tutor on user engagement and learning performance}, in: \bibinfo{booktitle}{International Conference on Social Robotics}, \bibinfo{organization}{Springer}. pp. \bibinfo{pages}{135--145}.
\bibitem[{Vygotsky(1978)}]{vygotsky1978mind}
\bibinfo{author}{Vygotsky, L.S.}, \bibinfo{year}{1978}.
\newblock \bibinfo{title}{Mind in society: The development of higher psychological processes}. volume~\bibinfo{volume}{86}.
\newblock \bibinfo{publisher}{Harvard university press}.
\bibitem[{Wang et~al.(2024)Wang, Yukiko and Chang}]{wang2024development}
\bibinfo{author}{Wang, X.}, \bibinfo{author}{Yukiko, M.}, \bibinfo{author}{Chang, H.H.}, \bibinfo{year}{2024}.
\newblock \bibinfo{title}{Development and techniques in learner model in adaptive e-learning system: A systematic review}.
\newblock \bibinfo{journal}{Computers \& Education} , \bibinfo{pages}{105184}.
\bibitem[{de~Wit et~al.(2018)de~Wit, Schodde, Willemsen, Bergmann, De~Haas, Kopp, Krahmer and Vogt}]{de2018effect}
\bibinfo{author}{de~Wit, J.}, \bibinfo{author}{Schodde, T.}, \bibinfo{author}{Willemsen, B.}, \bibinfo{author}{Bergmann, K.}, \bibinfo{author}{De~Haas, M.}, \bibinfo{author}{Kopp, S.}, \bibinfo{author}{Krahmer, E.}, \bibinfo{author}{Vogt, P.}, \bibinfo{year}{2018}.
\newblock \bibinfo{title}{The effect of a robot's gestures and adaptive tutoring on children's acquisition of second language vocabularies}, in: \bibinfo{booktitle}{Proceedings of the 2018 ACM/IEEE international conference on human-robot interaction}, pp. \bibinfo{pages}{50--58}.
\bibitem[{Yadollahi et~al.(2018)Yadollahi, Johal, Paiva and Dillenbourg}]{yadollahi2018deictic}
\bibinfo{author}{Yadollahi, E.}, \bibinfo{author}{Johal, W.}, \bibinfo{author}{Paiva, A.}, \bibinfo{author}{Dillenbourg, P.}, \bibinfo{year}{2018}.
\newblock \bibinfo{title}{When deictic gestures in a robot can harm child-robot collaboration}, in: \bibinfo{booktitle}{Proceedings of the 17th ACM conference on interaction design and children}, pp. \bibinfo{pages}{195--206}.
\bibitem[{Zaga et~al.(2015)Zaga, Lohse, Truong and Evers}]{zaga2015effect}
\bibinfo{author}{Zaga, C.}, \bibinfo{author}{Lohse, M.}, \bibinfo{author}{Truong, K.P.}, \bibinfo{author}{Evers, V.}, \bibinfo{year}{2015}.
\newblock \bibinfo{title}{The effect of a robot’s social character on children’s task engagement: Peer versus tutor}, in: \bibinfo{booktitle}{Social Robotics: 7th International Conference, ICSR 2015, Paris, France, October 26-30, 2015, Proceedings 7}, \bibinfo{organization}{Springer}. pp. \bibinfo{pages}{704--713}.

\end{thebibliography}

\end{document}